\documentclass[sigconf, 10pt]{acmart}

\AtBeginDocument{%
  }

\setcopyright{none} 

\acmConference[RobustGANTT]{}{Pre-Print}{Under Review.}
\renewcommand\footnotetextcopyrightpermission[1]{} 

\usepackage[utf8]{inputenc}
\usepackage[binary-units]{siunitx}
\usepackage{balance}
\usepackage{nicefrac} 
\usepackage{acronym}
\usepackage{xspace}
\usepackage{MnSymbol}

\usepackage{caption}
\usepackage{subcaption}
\usepackage{multirow}
\usepackage{booktabs}

\usepackage{enumitem}

\usepackage{stmaryrd} 

\newcommand{\fakeparagraph}[1]{\vspace{.5mm}\textbf{#1.}}

\newcommand{\fakepar}[1]{\fakeparagraph{#1}}

\newcommand{\oldsys}{DeepGANTT\xspace}
\newcommand{\system}{{RobustGANTT}\xspace}

\DeclareMathOperator{\diag}{diag}

\begin{document}

\title{Robust Generalization of Graph Neural Networks for Carrier Scheduling}


\author{Daniel F. Perez-Ramirez}
\orcid{0000-0002-1322-4367}
\affiliation{%
  \institution{RISE Computer Science \& \\ KTH Royal Institute of Technology}
  \country{Sweden}
}
\email{daniel.perez@ri.se}

\author{Carlos Pérez-Penichet}
\orcid{0000-0002-1903-4679}
\affiliation{%
  \institution{RISE Computer Science}
  \country{Sweden}
}
\email{carlos.penichet@ri.se}

\author{Nicolas Tsiftes}
\orcid{0000-0003-3139-2564}
\affiliation{%
  \institution{RISE Computer Science \& \\Digital Futures}
  \country{Sweden}
}
\email{nicolas.tsiftes@ri.se}

\author{Dejan Kostić}
\orcid{0000-0002-1256-1070}
\affiliation{%
  \institution{KTH Royal Institute of Technology \& RISE Computer Science}
  \country{Sweden}
}
\email{dmk@kth.se}

\author{Magnus Boman}
\orcid{0000-0001-7949-1815}
\affiliation{%
  \institution{Karolinska Institutet \& MedTechLabs}
  \country{Sweden}
}
\email{magnus.boman@ki.se}

\author{Thiemo Voigt}
\orcid{0000-0002-2586-8573}
\affiliation{%
  \institution{Uppsala University \& \\RISE Computer Science}
  \country{Sweden}
}
\email{thiemo.voigt@angstrom.uu.se}

\renewcommand{\shortauthors}{Perez-Ramirez et al.}

\begin{abstract}
    \emph{Battery-free sensor tags} are devices that leverage backscatter techniques to communicate with standard IoT devices, thereby augmenting a network's sensing capabilities in a scalable way.
    For communicating, a sensor tag relies on an unmodulated carrier provided by a neighboring IoT device, with a schedule coordinating this provisioning across the network.
    Carrier scheduling---computing schedules to interrogate all sensor tags while minimizing energy, spectrum utilization, and latency---is an NP-Hard optimization problem. 
    Recent work introduces learning-based schedulers that achieve resource savings over a carefully-crafted heuristic, generalizing to networks of up to 60 nodes. 
    However, we find that their advantage diminishes in networks with hundreds of nodes, and degrades further in larger setups. 
    This paper introduces \textbf{\system}, a GNN-based scheduler that improves generalization (without re-training) to networks up to 1000 nodes ($\mathbf{100}\boldsymbol{\times}$ training topology sizes). 
    \system not only achieves better and more consistent generalization, but also computes schedules requiring up to $\mathbf{2}\boldsymbol{\times}$ less resources than existing systems. Our scheduler exhibits average runtimes of hundreds of milliseconds, allowing it to react fast to changing network conditions. 
    Our work not only improves resource utilization in large-scale backscatter networks, but also offers valuable insights in learning-based scheduling.
\end{abstract}



\maketitle

\section{Introduction}
\begin{figure*}
    \centering
    \includegraphics[width=0.95\textwidth]{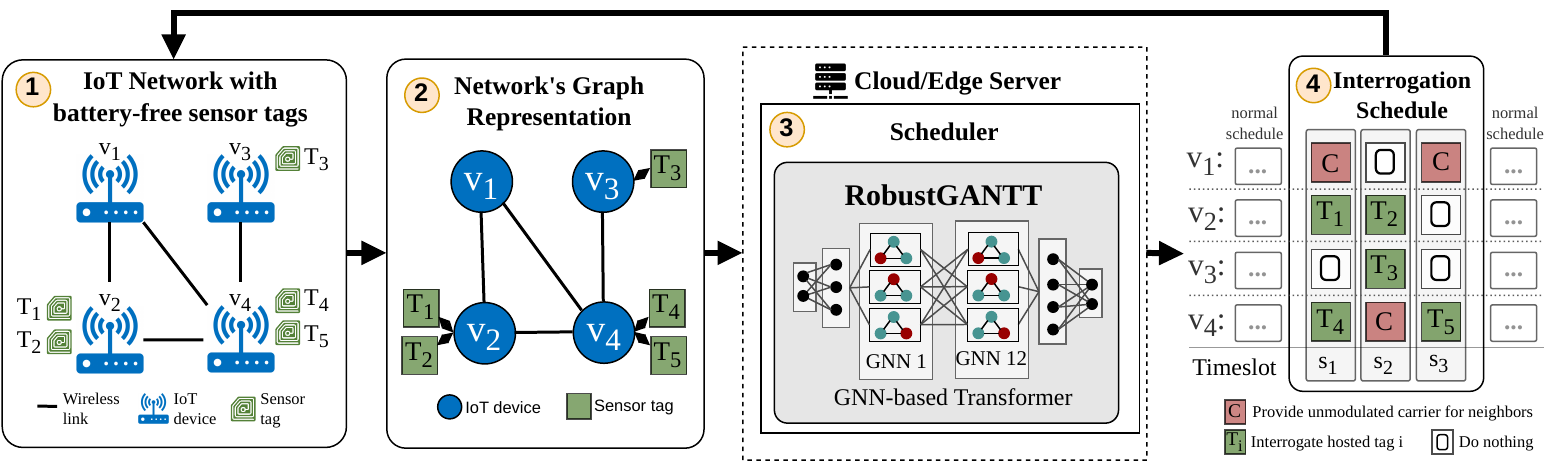}
    \caption{
        \textit{\system generates schedules for backscatter networks using a GNN-based Transformer model.}
        Step 1: collect MAC and routing protocol information. Step 2: build the IoT network's graph representation, only including edges strong enough for carrier provisioning (e.g., -75 dBm). Step 3: generate the schedule through iterative one-shot node classification. Step 4: disseminate the schedule using existing network flooding mechanisms and append it to the  IoT device's normal schedule.}
    \label{fig:graphical_abstract}
\end{figure*}

Recent advancements in backscatter communication enable the battery-free operation of sensor devices---termed \emph{sensor tags}---that perform bi-directional communication with standard \ac{IoT} devices~\cite{kellogg_passive_2016,ensworth_every_2015,kellogg2014wi,talla2017lora,iyer_inter-technology_2016,Perez-Penichet2016augmenting}.
Such sensor tags can be added to an existing network of \ac{COTS} \ac{IoT} devices to augment the network's sensing capabilities without requiring additional modifications to the \ac{IoT} devices~\cite{perez-penichet_tagalong_2020}.
However, communication between a sensor tag and its hosting \ac{IoT} device requires the provision of an unmodulated carrier by a neighboring \ac{IoT} device.
A schedule coordinates this provisioning globally across the network to interrogate all sensor values.
Figure~\ref{fig:graphical_abstract} shows the high-level procedure of computing a schedule, and its structure. It consists of one or more timeslots $s$, each assigning one of three possible actions to the \ac{IoT} devices in the network: provide unmodulated carrier $\mathtt{C}$, interrogate one of its hosted tags $\mathtt{T}$, or remain idle $\mathtt{O}$. 

\fakeparagraph{Motivation}
Battery-free sensor tags provide a scalable, cost- and energy-efficient way to augment the sensing capabilities of existing \ac{IoT} networks~\cite{kellogg_passive_2016, Perez-Penichet2016augmenting,perez-penichet_tagalong_2020}. 
Their battery-free operation reduces electronic waste, and prevents extensive maintenance costs compared to battery-powered alternatives.
It also allows placing sensors in hard-to-reach locations, such as medical implants, moving machinery, or embedded in physical infrastructure. The sensor tags may, e.g., prevent patients from undergoing surgery just to replace the battery of medical implants.
Reducing the energy consumption of networks hosting sensor tags is of paramount importance not only for sustainability reasons, but also because such networks are often energy constrained. 

\fakeparagraph{Challenges}
Carrier scheduling---computing a schedule to interrogate all sensor tags while minimizing energy, spectrum utilization, and latency---is, in general, an NP-Hard 
\ac{COP}~\cite{PerezPenichet2020afast}. 
It is similar to the traditional wireless link scheduling, but must consider additional constraints for tag interrogations and resource minimization (see Sec.~\ref{subsec:sched_constraints}).
There are also several symmetries involved, both in permuting the timeslots and in selecting carrier generators~\cite{perezramirez_DeepGANTT_2023}. E.g., in Figure~\ref{fig:graphical_abstract}, exchanging the timeslots' order alters neither the number of carriers required, nor the latency to query all tags. Also, for timeslot $s_3$, nodes $v_2$ and $v_3$ are equally valid carrier providers for $\mathtt{T}_5$.
A scheduler must process variable input-output structures: networks of different sizes, and schedules of different lengths. 
It must also leverage the topological structure of the network to favor using one carrier for multiple concurrent tag interrogations (e.g., timeslots $s_1$ and $s_2$ in Figure~\ref{fig:graphical_abstract}).
Additionally, it must compute schedules in a timely manner to react to connectivity changes of the \ac{IoT} network.

\fakeparagraph{Current Learning-based Schedulers exhibit Limited Scalability} In general, it is impractical to compute the analytically optimal schedule for \ac{IoT} networks of hundreds of nodes and sensor tags. 
This implies running a \ac{CO} for several hours, most likely yielding an obsolete schedule due to changes in the network's connectivity. 
Alternatively, one can use the TagAlong scheduler~\cite{PerezPenichet2020afast}, a carefully-crafted heuristic with polynomial runtime. However, its performance is increasingly sub-optimal as the network size increases. Recent work introduces \oldsys, a scheduler that learns from optimal schedules of small networks (up to 10 nodes) and scales to networks of up to 60 nodes, while reducing the number of carriers compared to TagAlong~\cite{perezramirez_DeepGANTT_2023}.
As we show in Sec.~\ref{subsec:perf_metrics_testbed}, reducing the number of carriers directly translates to energy savings.

\begin{figure}
    \centering
    \includegraphics[width=0.48\textwidth]{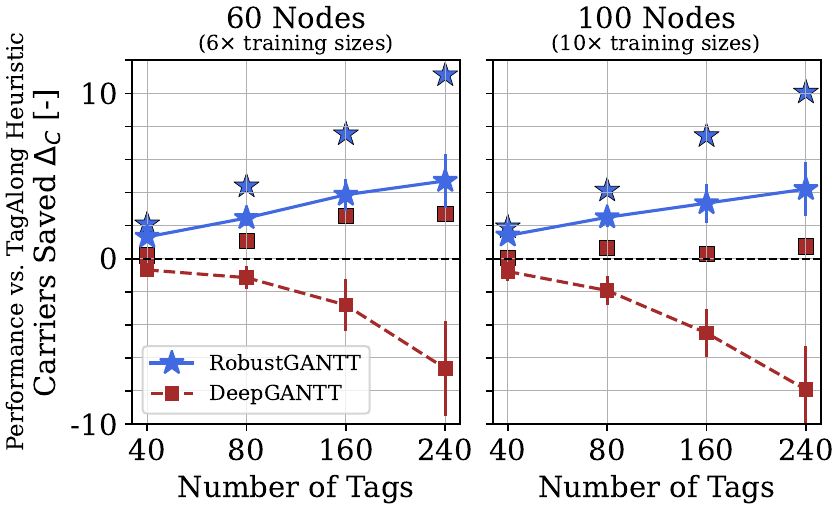}
    \caption{
        \textit{\system has better and more consistent generalization to larger topologies (higher is better).}
        We train eight identical scheduler models for both \system (stars) and \oldsys~\cite{perezramirez_DeepGANTT_2023} (squares), and compare them against the TagAlong heuristic~\cite{PerezPenichet2020afast} on larger, previously unseen topologies (without re-training).
        Isolated markers depict the best performing model, markers joined by lines represent the average, and vertical lines depict standard error.} 
    \label{fig:intro_plot}
\end{figure}

However, \oldsys presents two main issues when further scaling the problem to graphs larger than 60 nodes, as depicted in Figure~\ref{fig:intro_plot}.
We train eight independent models (in accordance to ~\cite{perezramirez_DeepGANTT_2023}), while fixing the training data, hyperparameters and random seeds.
\oldsys's best model (isolated squares in Figure~\ref{fig:intro_plot}) is only marginally better than the heuristic for 100-node topologies. 
Moreover, while all eight models perform well on the training set, their generalization to larger networks significantly varies. 
The dashed line in Figure~\ref{fig:intro_plot} shows how the average performance across the eight models is increasingly worse compared to TagAlong, even for 60-node topologies.
We attribute this behavior both to the stochastic training procedure that leads most scheduler models to "bad" local minima, and to the model's inability to capture the full problem complexity. 

\fakeparagraph{Approach}
In this paper, we leverage the latest advances in \acp{GNN} and \ac{ML} to present \system, a scheduler for backscatter networks with strong and consistent generalization capabilities.
To design \system, we set out to explore \ac{ML}-related training aspects, beginning with our own implementation of \oldsys. We train our scheduler with optimal schedules of networks of up to 10 nodes and 14 tags computed by a \ac{CO}. 
The use of \acp{GNN} in our system design allows the scheduler to process variable input-output structures, and to process much larger, previously unseen topologies without the need for re-training.
For designing our system, we investigate three aspects influencing the scheduler's generalization as follows. 

First, we assess the influence of warmup~\cite{ma2021adequacy}, and prove it highly beneficial for the model's ability to compute complete schedules for larger topologies. 
Furthermore, we explore incorporating \ac{PE} into the node features to enhance the \ac{GNN}'s ability to handle symmetries in schedule computation.
We find that the node-degree \ac{PE} offers the best trade-off for achieving good generalization, while avoiding the computation overhead of \ac{EVD}-based methods.
Finally, we study the influence of increasing the number of attention heads of the \ac{GNN} layers to capture more complex topological dependencies among the \ac{IoT} nodes in the network~\cite{Nakkiran2020DeepDD}.

\fakeparagraph{Contributions}
Based on the former, we present \system, a novel \ac{GNN}-based scheduler that generalizes to networks of up to 1000 nodes ($100\times$ training sizes), far beyond the capabilities of current learning-based systems~\cite{perezramirez_DeepGANTT_2023}, while delivering schedules that require up to $2\times$ less resources than those by the TagAlong heuristic~\cite{PerezPenichet2020afast}. Our system exhibits polynomial time complexity, allowing it to react fast to changing network conditions. Figure~\ref{fig:intro_plot} shows how our scheduler not only outperforms \oldsys, but also exhibits consistent generalization across the independently trained models. 

To evaluate \system's capabilities on real-life \ac{IoT} networks, we use it to compute schedules for a testbed with 23 nodes and varying number of sensor tags. 
Our system achieves 12\% on average and up to 53\% savings in energy and spectrum utilization compared to the TagAlong heuristic, which corresponds to up to $1.9\times$ more savings over the \oldsys scheduler. 
Furthermore, thanks to the polynomial time complexity of the model, it exhibits average runtime of 540 ms for the real \ac{IoT} network, and achieves up to $2\times$ reduction in 95th percentile runtime against \oldsys.
These characteristics enable \system to compute schedules for \ac{IoT} networks even in dynamic changing conditions.

We make the following specific contributions:
\begin{itemize}
    \item We present \system, a learning-based scheduler that generalizes without re-training to networks of up to 1000 nodes ($\mathbf{100}\boldsymbol{\times}$ larger than those used for training), far surpassing existing learning-based schedulers.  
    \item We use \system to compute schedules for a real \ac{IoT} network. Our model achieves 12\% on average and up to 53\% resource savings compared to TagAlong, which correspond to up to $1.9\times$ more savings than those achieved by \oldsys.
    \item \system reduces runtime's 95th percentile by up to $2\times$ against \oldsys, which allows it to react faster to changing network conditions.  
\end{itemize}

The paper is structured as follows. Sec.~\ref{sec:background} provides background and related work. Sec.~\ref{subsec:optimizationproblem} formally describes the scheduling problem. Sec.~\ref{sec:robustgantt} presents the \system scheduler, and Sec.~\ref{sec:gnn_design} describes our system's GNN model design. Sec.~\ref{sec:evaluation} and Sec.~\ref{sec:discussion} present the evaluation and discussion, respectively. Finally, Sec.~\ref{sec:conclusion} concludes the paper.

\section{Background and Related Work}
\label{sec:background}

Our work draws upon backscatter communication, scheduling for backscatter networks and \ac{ML} for scheduling.

\subsection{Backscatter Communication}
\label{subsed:tagAugmentedNetworks}
Several recent efforts advance backscatter communications and battery-free
networks~\cite{kellogg_passive_2016,talla2017lora,iyer_inter-technology_2016,ensworth_every_2015,kellogg2014wi,zhang2017freerider,jansen_multihopbackscatter_2019,karimi_design_2017,geissdoerfer_bootstrapping_2021, ahmad2021enabling, li2018passive, guo2020aloba}.
While some work focus on monostatic or multi-static backscatter configuration~\cite{liu_season_2011, hamouda_reader_2011, chen_time-efficient_2012, katanbaf2021multiscatter}, we focus on networks hosting sensor tags in the bistatic configuration (separated receiver from carrier generator).

\begin{figure}
    \centering
    \includegraphics[width=0.48\textwidth]{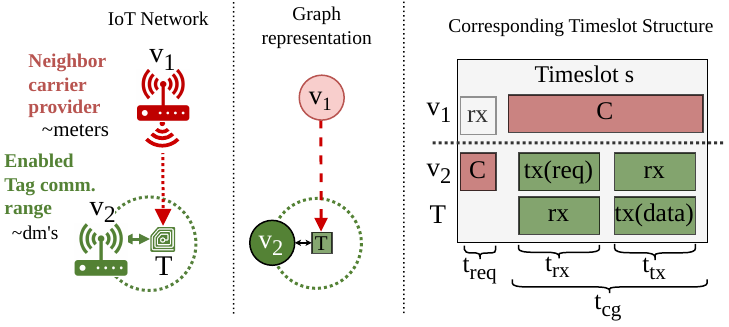}
    \caption{Backscatter communication between a tag $\mathtt{T}$ and its host $v_2$ when assisted by neighboring node $v_1$ during a timeslot $s$. 
    }
    \label{fig:tag-to-host-comm}
\end{figure}

Sensor tags leverage backscatter techniques to perform bidirectional communication with their hosting \ac{IoT} node over standard physical layer protocols~\cite{kellogg2014wi,ensworth_every_2015,Perez-Penichet2016augmenting,talla2017lora}.
They achieve their low-power operation by offloading the local oscillator to a neighboring \ac{IoT} node (different from its host), which provides the tag with an unmodulated carrier~\cite{perez-penichet_tagalong_2020}. 
The \ac{COTS} \ac{IoT} nodes achieve this by, e.g., using their radio test mode~\cite{Perez-Penichet2016augmenting}. 
An \ac{IoT} node in the network hosts zero or more sensor tags. Moreover, we assume that a sensor tag is hosted by \textit{exactly} one \ac{IoT} node responsible for querying the sensor readings. 
Sensor tags are located within decimeters range to its hosting \ac{IoT} node, while the \ac{IoT} nodes in the network are within meters from each other (see Figure~\ref{fig:tag-to-host-comm}).

\fakepar{Node-to-Tag Communication}
The host-to-tag communication occurs over a time-slotted channel access mechanism due to its ease of integration of sensor tags and their widespread use in commodity devices. Both Bluetooth and Zigbee/IEEE 802.15.4 support this in their standards~\cite{bluetooth_2021, ieee802_15_4_2016}.
Figure~\ref{fig:tag-to-host-comm} describes the communication between a tag $\mathtt{T}$ and its host $v_2$, when assisted by a neighboring carrier provider \ac{IoT} node $v_1$. $t_{rx}$ and $t_{tx}$ are the times for the sensor tag to receive the request-to-transmit from its host, and for transmitting the sensor value back, respectively. $t_{cg}$ is the time spent in carrier provisioning for tag-to-host communication.
The timeslot is long enough to complete one request-response cycle between a node and a tag---e.g., two consecutive \ac{TSCH} timeslots (10 ms each) for both transmitting the request to the tag and receive the response~\cite{perez-penichet_tagalong_2020,perezramirez_DeepGANTT_2023}. 
During $t_{req}$, $v_2$ sends a request signal to $v_1$ to start carrier provisioning, allowing $v_2$ to regulate the frequency of tag interrogation---e.g., in a schedule with 10 timeslots (200 ms total duration with \ac{TSCH}), a node might not want to query its tag $1000 \text{ ms} / 200 \text{ ms} = 4$ times per second.

\begin{figure}
    \centering
    \subfloat[\footnotesize Optimal schedules leverage the topology structure to assign carriers such as to minimize $C$ and $L$. In general, it holds $C\geq L$.\label{subfig:schedule-types}]{%
       \includegraphics[width=0.45\textwidth]{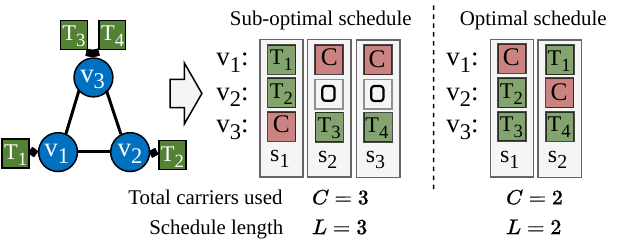}}
    \\
    \subfloat[Efficient carrier re-use.
    \label{subfig:carrier-reuse}]{%
        \includegraphics[width=0.235\textwidth]{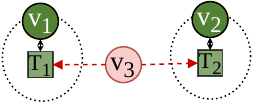}}
    \hfill
    \subfloat[Carrier interference.
    \label{subfig:carrier-interference}]{%
        \includegraphics[width=0.235\textwidth]{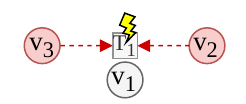}}
       
    \caption{Example of schedules for a network topology and efficient carrier re-use for concurrent tag interrogation.} \label{fig:schedules}
\end{figure}

\fakeparagraph{Schedule}
\label{subsec:sched_constraints}
A schedule coordinates the interrogation of all sensor tags and the provisioning of unmodulated carriers by the \ac{IoT} nodes for such purposes. 
It consists of $L\geq1$ \emph{timeslots}, each assigning one of three possible actions to \ac{IoT} nodes in the network: interrogate one of its tags $\mathtt{T}$, provide unmodulated carrier $\mathtt{C}$ for neighboring tags, or remain idle $\mathtt{O}$. 
We leverage the spatial distribution of nodes and tags to perform concurrent tag interrogations with one carrier provider (see Figure~\ref{subfig:carrier-reuse}).
There are two constraints for performing tag interrogations~\cite{perez-penichet_tagalong_2020}. First, due to the time-slotted channel access control mechanism, a node can interrogate only one of its hosting tags per timeslot. 
Additionally, for a tag to communicate with its hosting node, exactly one neighboring \ac{IoT} node must provide it with an unmodulated carrier. Multiple impinging carriers on a sensor tag causes interference, and prevents proper tag interrogations (see Figure~\ref{subfig:carrier-interference}).

\fakepar{Resource Efficiency}
\label{subsec:reseff_NPhard}
Two metrics determine a schedule's resource efficiency: the length of the schedule $L$ and the number of carrier slots $C$. 
While $L$ indicates the latency of querying all sensor values, $C$ is directly related to spectrum utilization and energy consumption of the \ac{IoT} network (see Sec.~\ref{subsec:perf_metrics_testbed}). 
Figures~\ref{subfig:schedule-types} and~\ref{subfig:carrier-reuse} show how resource efficient schedules exploit the topological structure of the network to re-use carrier generating nodes within a timeslot.

\subsection{Existing Schedulers}

A scheduler is a system that receives a description of the \ac{IoT} network hosting sensor tags, and computes a schedule for interrogating the sensor tags. 
While recent work explores autonomous scheduling for TDMA based networks~\cite{duquennoy2015orchestra}, carrier scheduling requires more powerful hardware for such purposes.
In general, carrier scheduling can be solved analytically by using a \ac{CO} to obtain the optimal schedule. However, this is only feasible for small-sized \ac{IoT} networks, since the NP-Hard nature of the problem prevents the practical application of the \ac{CO} due to the long runtimes (e.g., up to 10 hours for a 10-node network). 

Alternatively, Pérez-Penichet et al. present TagAlong~\cite{PerezPenichet2020afast}, a heuristic algorithm that uses graph coloring to compute schedules. While TagAlong exhibits polynomial runtime, its performance becomes increasingly sub-optimal as the network size increases. Additionally, Pérez-Ramírez et al. present \oldsys~\cite{perezramirez_DeepGANTT_2023}, the first \ac{ML}-based system for carrier scheduling that iteratively builds the schedule timeslot by timeslot. 
\oldsys learns from optimal schedules (computed by a \ac{CO}) of networks of up to 10 \ac{IoT} nodes and 14 sensor tags~\cite{perezramirez_DeepGANTT_2023}. It generalizes to networks of up to 60 nodes, achieving significant reduction in the number of carriers required in the schedule against TagAlong. In this work, we advance learning-based scheduling by considering networks of hundreds of nodes, far beyond \oldsys's capabilities.

\subsection{Learning-based Scheduling}

Several works explore applying \ac{ML} and \acp{GNN} for both \ac{COP} and scheduling~\cite{Vinyals2015pointernets, vesselinova2020learning, Bengio2021mlforcop, Dai2017learningcops, Li2018copsgcn, Manchanda2020learning, jeon2022neural, mao2019learning}, but few explore their usage for backscatter networks~\cite{perezramirez_DeepGANTT_2023}.
In this work, we explore \acp{GNN} to design a system that generates schedules for backscatter networks consisting of hundreds of nodes. 

\fakeparagraph{Graph Neural Networks} 
\acp{GNN} are a flexible \ac{ML} tool for tackling various inference tasks on graphs, such as node classification~\cite{Scarselli2009gnnmodel, hamilton2020graph, zonghan2021gnnsurvey}. 
Intuitively, stacking $K$ \ac{GNN} layers generates node embedding vectors that consider their $K$-hop neighborhood by utilizing the graph's structure and the relationships between nodes~\cite{Gilmer2017mpnn, Kipf2017gcn}.
These embeddings are generally processed further with linear layers to produce the final output based on the specific task. For instance, node classification can be achieved by feeding each node embedding vector through a classification layer.
For a graph $G=\langle V,E \rangle$ with nodes $v \in V$ and edges $(v, u) \in E$, at \ac{GNN} layer $i$, each node feature vector $h_v$ is updated as:
\begin{equation}
    h^{i}_v = f_1\left(\,\,h^{i-1}_v,\,\, \bigcup_{u\in \mathcal{N}(v)}\left[ f_2\left(h_u^{i-1}\right) \right] \,\,\right) \, \text{,}
    \label{eq:message-pas}
\end{equation}
where $\mathcal{N}(v)$ is the set of neighbors of node $v$ with $h_u$ representing their feature vectors, and $\bigcup$ is a commutative aggregation function. $f_1, f_2$ are non-linear transformations~\cite{Gilmer2017mpnn}. 
For attention-based \acp{GNN}, additional learnable scaling parameters are included within $\bigcup$ to weight the contributions of neighboring nodes differently. 

One key advantage of \acp{GNN} is their ability to leverage the structural dependencies within the graph, and their ability to perform inference on new graphs not encountered during training without needing to retrain the model~\cite{Hamilton2017inductive, Velickovic2018gat, vesselinova2020learning}.

\fakeparagraph{\ac{PE} in \acp{GNN}} 
\ac{PE} augments each node's input feature vector with additional information of its structural role in the graph. The intuition is to aid subsequent \ac{GNN} layers to better distinguish the nodes involved in symmetries---i.e., to perform injective aggregation of neighboring nodes' features.
Recent work explore \ac{PE} with both local and global graph properties~\cite{wang2022equivariant,belkin2003laplacian,Diwedi2023benchmarking,lim2022sign, huang2024spe, rampavsek2022recipe}. While most focus on using \ac{PE} to better distinguish different graphs, we are interested in assessing their advantage for effective node classification.

\section{Carrier Scheduling Problem}
\label{subsec:optimizationproblem}
The \ac{COP} of computing a schedule to interrogate all sensor tags in an \ac{IoT} network while minimizing both the length of the schedule $L$ and the number of carrier slots $C$ is described as follows.
We model the network as an undirected connected graph $G$, defined by the tuple $G=\langle V_a,E \rangle$, where $V_a$ is the set of $N$ \ac{IoT} nodes in the network $V_a=\{v_i\}_{i=0}^{N-1}$, and $E$ is the set of edges between the nodes $E=\{\langle u,v\rangle|u,v \in V_a\}$. 
The connectivity among \ac{IoT} nodes is determined by the
wireless link signal strength, i.e., there
is an edge between two nodes only if there is a sufficiently strong
wireless signal for providing unmodulated
carrier~\cite{PerezPenichet2020afast, perezramirez_DeepGANTT_2023}.
We denote the set of $T$ tags in the network as $N_t=\{t_i\}_{i=1}^{T}$, and their respective tag-to-host assignment as $H_t: t\in N_t\mapsto v\in V_a$. 
The role of a node $v$ within a timeslot $s$ 
is indicated by the map $R_{v,s}: v\!\in \!V_a, s\!\in \![1, L]
\mapsto \{\mathtt{C}, \mathtt{T}, \mathtt{O}\}$, where $L$ is the schedule
length in timeslots. Hence, a timeslot $s_j$ consists of an $N$-dimensional
vector containing the roles assigned to every node during timeslot $j$:
$s_j=\left[R_{v_i,j} | v_i\in V_a \right]^{\top}$.

For a given problem instance $g=\langle G, N_t, H_t\rangle$, the carrier scheduling problem is formulated as follows:
\begin{eqnarray}
\min\,&{}&\,\left(T \cdot  C + L \right) \label{eq:optobj}\\
\text{s.t.}\,&{}& \,\forall t\!\in\!N_t \,\,\exists!\,\,s\!\in\![1, L]:\, R_{H_t,s}=\mathtt{T}\label{eq:optconst1}\\ \nonumber 
\,&{}& \,\forall s\!\in\![1, L]\,\,\forall t\!\in\!N_t\,|\,R_{H_t,s}=\mathtt{T}  
\; 
\exists!\,v_j\!\in\!V_a: \\
&{}& \, R_{v_j,s}=\mathtt{C} \wedge (H_t, v_j)\in E \,\,\text{,} \label{eq:optconst2}
\end{eqnarray}
where $C$ is the total number of carriers required in the schedule. 
Constraints (\ref{eq:optconst1})
and (\ref{eq:optconst2}) enforce that tags are interrogated only once in the
schedule and that there is exactly one carrier-providing neighbor per tag in each
timeslot (to prevent collisions), respectively.
The objective function (\ref{eq:optobj}) prioritizes reducing $C$ over $L$ because we are most concerned with energy and spectrum efficiency---reducing $C$ often implies a reduction of $L$, but the converse is not
necessarily true~\cite{perezramirez_DeepGANTT_2023}. 

\fakepar{Symmetry-Breaking Constraints}
\label{subsec:symmetry_breaking}
Solutions to the carrier scheduling problem are highly symmetrical, which limits effective training of a supervised \ac{ML} model~\cite{perezramirez_DeepGANTT_2023}.
Symmetries arise both from the network topology and from the sensor tags' distribution among the nodes. E.g., for a star topology hosting one sensor tag in the center node, any of the leaf nodes can be the carrier provider, but the scheduler needs to select only one of these. 
Additionally, we do not assume any a-priori order for tag interrogations. Hence, any of the $L!$ permutations of a schedule's timeslots is also a valid schedule with the same length $L$ and number of carrier slots $C$.

Symmetry-breaking constraints allow to efficiently learn the behavior of the optimal scheduler and properly train an \ac{ML} model~\cite{perezramirez_DeepGANTT_2023}. For the training data generation procedure, we further constrain the optimization objective in Eq.~\ref{eq:optobj} by enforcing two lexicographical minimizations: first of a vector of length $T$ (number of tags) that indicates the timeslot when each tag is interrogated, and another length-$T$ vector containing the node that provides the carrier for each tag.

\section{\system System Design}
\label{sec:robustgantt}

We consider networks consisting of \ac{COTS} wireless \ac{IoT} devices, or \emph{nodes}, equipped with radio transceivers that support standard physical layer protocols, such as Bluetooth or IEEE 802.15.4/ZigBee. 
These nodes perform their regular computation and communication tasks according to their normal schedule~\cite{perez-penichet_tagalong_2020, duquennoy2015orchestra}. 
The \ac{IoT} nodes are either battery-powered or connected to mains power. 
We extend the sensing capabilities of the nodes with battery-free sensor tags~\cite{Perez-Penichet2016augmenting,perez-penichet_tagalong_2020}, which require an additional schedule to coordinate carrier provisioning and tag interrogations. This schedule is appended to the \ac{IoT} network's normal schedule. 

We base our system design on \oldsys and set to explore \ac{ML} related aspects to design a scheduler with better and more robust generalization to larger networks. 
\begin{figure*}
    \centering
    \includegraphics[width=0.85\textwidth, trim={0 0 3.2cm 0}, clip]{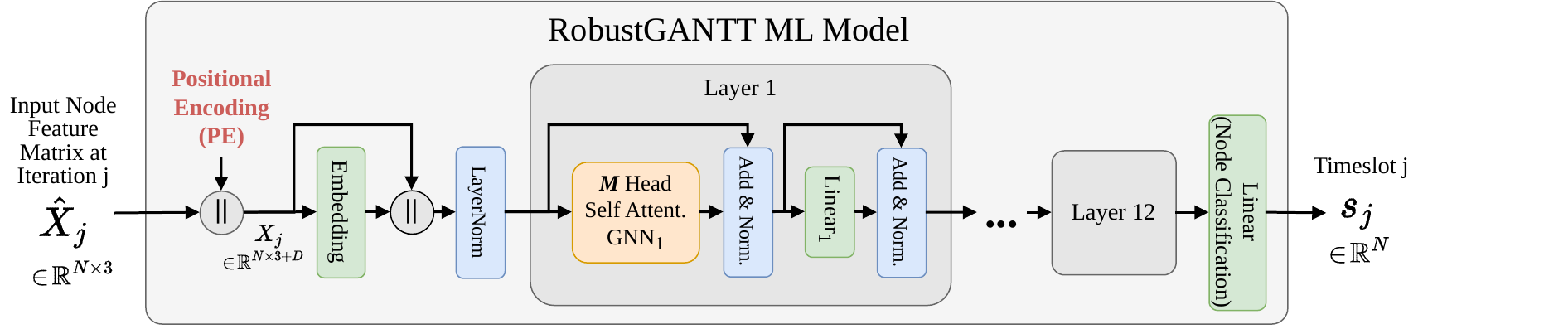}
    \caption{
        \textit{\system's \ac{ML} model architecture}. 
        It receives as input a node-feature matrix $\hat{X}_j$ and produces the corresponding schedule timeslot $s_j$ for every iteration $j$. There is a multi-head self-attention \ac{GNN} in each of \system's layers (orange box). $||$ represents a concatenation operation. Green boxes represent a non-linear transformation by a single-layer fully-connected neural network.
        }
    \label{fig:ml_model}
\end{figure*}

\subsection{System Description}

\system resides at the Edge/Cloud, and asynchronously receives requests by one or multiple \ac{IoT} networks hosting battery-free sensor tags to compute schedules. 
Note that this is also true for any scheduler to tackle this problem due to the computational demands required in computing schedules.
The interaction between \system and the \ac{IoT} network is depicted in Figure~\ref{fig:graphical_abstract}. 

First, the \ac{IoT} network collects the MAC and routing protocol information to build the network topology $G$ and the tag-to-host mapping $H_t$. In our evaluation in Sec.~\ref{sec:evaluation}, we use metrics from both \ac{TSCH}~\cite{duquennoy17tsch} and RPL~\cite{winter2012rpl}, but the process is analogous for other physical layer and routing protocols. 
Upon detection of changes either in the network's connectivity or in the tag-to-host mapping, the network issues a request to \system for computing a new schedule.
Next, the scheduler receives the network information $g$ and performs iterative node classification using a \ac{GNN} model to compute the interrogation schedule timeslot by timeslot. 
Finally, \system delivers the schedule back to the \ac{IoT} network, where it is disseminated using existing network flooding mechanisms, such as Glossy~\cite{ferrari2011glossy}. 

\label{subsec:gnnModel}
At the core of \system lies an attention-based \ac{GNN} model to perform iterative one-shot node classification. In each iteration $j$, the \ac{GNN} model receives as input a node feature matrix $X_j\in\mathbb{R}^{N\times D}$ with $D=3$ features per node, and delivers as output the scheduling timeslot $s_j\in\mathbb{R}^{N}$. The resulting $s_j$ corresponds to assigning each of the $N$ nodes to one of three possible classes $\{\mathtt{T}$, $\mathtt{C}$, $\mathtt{O}\}$.

\system keeps a cached representation of the topology $G$ and the tag-to-host mapping $H_t$ that is updated after each iteration.
After computing the $j^{\text{th}}$ timeslot $s_j$, the tags assigned to be interrogated are removed from the cached representation of the topology, and a new input feature matrix is generated $X_{j+1}$ to compute the next scheduling timeslot $s_{j+1}$. Being a probabilistic model, \system has a component for checking that $s_j$ complies with the scheduling constraints at each iteration. This process is repeated until there are no more tags in the cached topology.

\subsection{Scheduling Approach}
\fakepar{Input Node Feature Matrix} Upon receiving the \ac{IoT} network information, \system builds a graph representation of the topology and parses this information for input to the \ac{GNN} model.
The input node feature matrix to the \ac{GNN} $X_j$ consists of $D=3$ features per node: 
\begin{enumerate}[leftmargin=*]
    \item Hosted-Tags: the number of tags hosted by the node. \label{feat1}
    \item Node-ID: integer identifying the node in the graph.\label{feat2}
    \item Min. Tag-ID: integer that represents the minimum tag ID among tags hosted by the node.\label{feat3}
\end{enumerate}
Intuitively, Hosted-Tags is decisive for assigning carrier-generating nodes -- the node hosting the greatest number of tags in the network should avoid providing unmodulated carriers. 
Thanks to the symmetry-breaking constraints (\S\ref{subsec:optimizationproblem}), including features \ref{feat2} and \ref{feat3} provides the scheduler with context on how to prioritize carrier-provider nodes, and with an order to interrogate the tags, respectively. In practice, network operators can exploit this by, e.g., prioritizing \ac{IoT} nodes connected to mains power as carrier providers, or by prioritizing certain tags to be interrogated early in the schedule, simply by assigning them a lower node/tag-ID.

\fakepar{ML Model Architecture}
Figure~\ref{fig:ml_model} depicts the system's \ac{ML} model. The node feature matrix is first passed through a node-wise embedding layer, followed by a concatenation and layer normalization operation. Subsequently, the hidden representation is passed through a stack of 12 \ac{GNN} layers, each containing both a linear activation and self-attention \ac{GNN}. We fix 12 as the number of layers due to its wide application in language modelling with both GPT and BERT~\cite{devlin2018bert, radford2019gptlanguage,radford2018gptimproving}, and its success in learning-based schedulers~\cite{perezramirez_DeepGANTT_2023}.
The linear layer is a fully-connected neural network that acts on each node intermediate feature vector independently, while the \ac{GNN} uses a multi-head attention mechanism of $M$ heads for computing message passing operations~\cite{Velickovic2018gat}. The structure and skip connections of each \ac{GNN}-Block is inspired by the Transformer architecture~\cite{Vaswani2017attention}.

\subsection{Model Training}
\label{subsec:model-trainig}
We train \system with optimal schedules from relatively small networks that are computed by the optimal scheduler. We then use \system to compute schedules for much larger and previously-unseen networks without the need for the scheduler to be re-trained.

As loss function, we select the modified cross-entropy loss that includes both a scaling factor to give more importance to the carrier generator class $\mathtt{C}$~\cite{perezramirez_DeepGANTT_2023}, and L2 weight regularization~\cite{lecun1989optimal,krogh1991simple}. As optimizer, we use Adam with its default hyperparameters~\cite{Kingma2015adam}. We use learning rate decay by 2\% every epoch, with an initial learning rate $\epsilon_{init}\!=\!10^{-3}$. We early stop model training after 25 consecutive epochs without minimization of the validation loss, and save the best performing model based on the carrier-class F1-score~\cite{perezramirez_DeepGANTT_2023}. 

\section{System GNN Model Design}
\label{sec:gnn_design}
We explore \ac{ML}-related design aspects that provide \system with strong and consistent generalization to larger, previously unseen, \ac{IoT} networks. 
We believe our findings not only advance carrier scheduling, but also provide insights on designing learning-based schedulers for \ac{IoT} networks. 

\fakeparagraph{Setup}
We undergo a structured and sequential process in three stages, selecting the best configuration in each stage before transitioning to the next one: i) learning rate warmup, ii) local and global \ac{PE}, and iii) increasing the number of attention heads.
For each stage, we train multiple models according to Sec.~\ref{subsec:model-trainig} using the training dataset from Sec.~\ref{subsec:training_data}, while fixing the hyperparameter configuration. To mitigate the effect of randomness, we fix the random seeds from software libraries at the application level: Python, PyTorch, and NumPy~\cite{pytorch2019, pytorchReprod2024}.
Since the best performance for a given model configuration may greatly diverge from its average (see Figure~\ref{fig:intro_plot}), we train multiple, but identical, \ac{ML} models for each configuration to assess their performance consistency to larger topologies.
However, we are limited to training 4-8 models per configuration, since the training and subsequent deployment to larger graphs takes between 10-45 hours for a single model, depending on its configuration. 
Our analysis results in the training of over 50 \ac{ML} scheduler models. 

After training, we deploy the models to compute schedules for the generalization dataset -- previously unseen topologies of larger size than those trained (see Sec.~\ref{subsec:generalization_data}). No re-training is done at this stage.
We report mean and percentile statistics across the runs for each model configuration, and select the best one based on the performance metrics from Sec.~\ref{subsec:perf_metrics}.

We highlight the following \textbf{key findings}: 
\begin{itemize}[leftmargin=*]
    \item Warm-up significantly contributes to computing complete and correct schedules for larger topologies.  
    \item Node degree \ac{PE} allows for a good trade-off to assist in breaking graph symmetries with a low-overhead \ac{PE} method. 
    \item 12 attention heads consistently achieves good generalization performance to larger topologies. 
\end{itemize}

\subsection{Datasets}
\label{subsec:dataset}
We train all models using the data fom Sec.~\ref{subsec:training_data}. After training, their performance is compared on the dataset described in Sec.~\ref{subsec:generalization_data}, on which the models are \textit{not} trained.

\subsubsection{Training Dataset}
\label{subsec:training_data}
We use artificially generated problem instances (topologies and tag assignments) according to Perez-Ramirez et al.~\cite{perezramirez_DeepGANTT_2023}. The dataset contains 580000 problem instances with networks of 2-10 nodes and 1-14 tags that are randomly assigned. We use the \textit{optimal scheduler} to obtain schedules for these problem instances. This implies using a \ac{CO} to solve analytically the \ac{COP} described in Sec.~\ref{subsec:optimizationproblem}. We use 80\%-20\% training and validation data splits.

\subsubsection{Generalization Dataset}
\label{subsec:generalization_data}
Consists of larger and previously unseen topologies on which models are not trained. We select the best performing model configuration in this dataset when deciding the final \ac{ML} model. We consider $200$ problem instances (network topologies and tag assignments) for every $(N, T)$ pair from the sets $N\in\{10, 20, 40, 60, 80, 100\}$ nodes and $T\in\{40, 80, 160, 240\}$ tags---i.e., $4800$ networks. 

\subsection{Performance Metrics}
\label{subsec:perf_metrics}
In this work, we are interested in the system-related aspects of \system. Hence, we consider the following application-related performance metrics in \ac{ML} model design. 

\fakepar{$\mathbf{\Pi}$---Correctly Computed Schedules} 
Given a set of \ac{IoT} networks, $\Pi\in[0, 100]\%$ represents the percentage of networks for which \system produces a complete schedule -- one that interrogates all sensor tags. 
Since \system is a probabilistic model, we must account for cases in which the scheduler cannot produce all the required timeslots to query all sensor values in the network. 
If \system fails to deliver \emph{all} timeslots, even if it correctly delivered some of them, we consider it a failed schedule. 

\fakepar{$\boldsymbol{\Delta}_\mathbf{C}$---Carriers Saved} This metric directly relates to the energy and spectrum utilization of the network. It compares the total number of carrier generator slots $C$ from the schedule generated by the TagAlong heuristic $C_{ta}$ against the total number of carrier slots from the schedule computed by a learning-based scheduler $C_{nn}$ as: $\boldsymbol{\Delta}_\mathbf{C}= C_{nn} - C_{ta}$.


\subsection{Results}
\label{subsec:results}

We describe the considered \ac{ML} design aspects and their influence in our system's generalization to larger topologies.

\subsubsection{Influence of Warmup}
\label{subsec:warm-up}
Based on the findings from Ma et al. \cite{ma2021adequacy}, we evaluate the influence of learning rate warm-up on the optimization. It involves starting training with a small learning rate $\Tilde{\epsilon}\ll\epsilon_{init}$ and gradually increase $\Tilde{\epsilon}$ until reaching the initial learning rate $\epsilon_{init}$. 
Intuitively, warmup provides more stability by regularizing the magnitude of parameter updates in early stages of training for momentum-based optimizers. Since such optimizers perform the parameter updates considering past statistical moments of the gradients, warmup allows the optimizer to calculate moments' statistics before performing big jumps in the parameter update, which reduces variance of the update steps~\cite{ma2021adequacy}.

We choose an untunned linear warmup schedule~\cite{ma2021adequacy} due to its simplicity and competitive performance. It requires $2*(1-\beta_2)^{-1}$ steps so that $\Tilde{\epsilon}\approx\epsilon_{init}$, where $\beta_2=0.999$ is Adam's second moment decay rate~\cite{Kingma2015adam}. The warm-up update of the learning rate is performed as:
$
    \Tilde{\epsilon} = \epsilon_{init}*\min\left(1, \frac{1-\beta_2}{2}*i\right) \text{,}
$
where $i$ is the mini-batch iteration.
We independently train two sets of eight identical models, with and without warmup.

\begin{figure} 
    \centering
  \subfloat[\textbf{\oldsys} ML model (\textbf{no} warm-up). \label{subfig:nowarmup}]{%
       \includegraphics[width=0.475\textwidth]{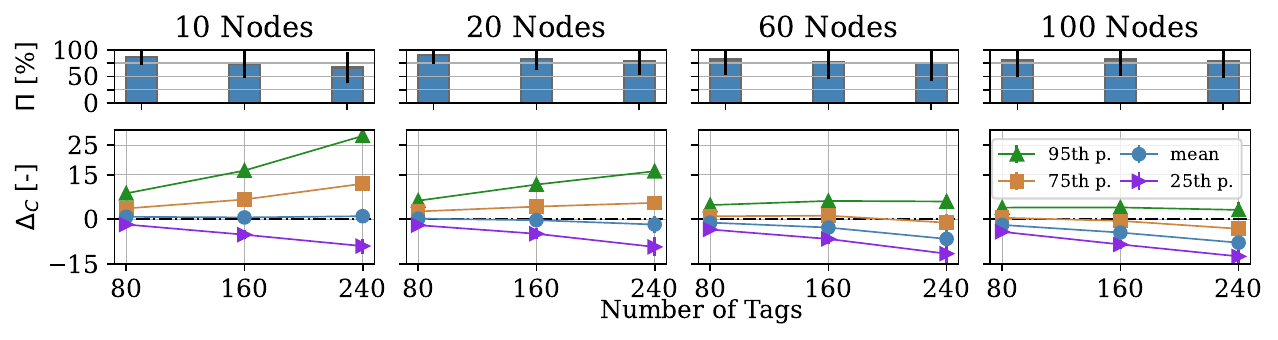}}
    \\
  \subfloat[2 Heads \textbf{with warm-up}. 
  \label{subfig:2H}]{%
        \includegraphics[width=0.475\textwidth]{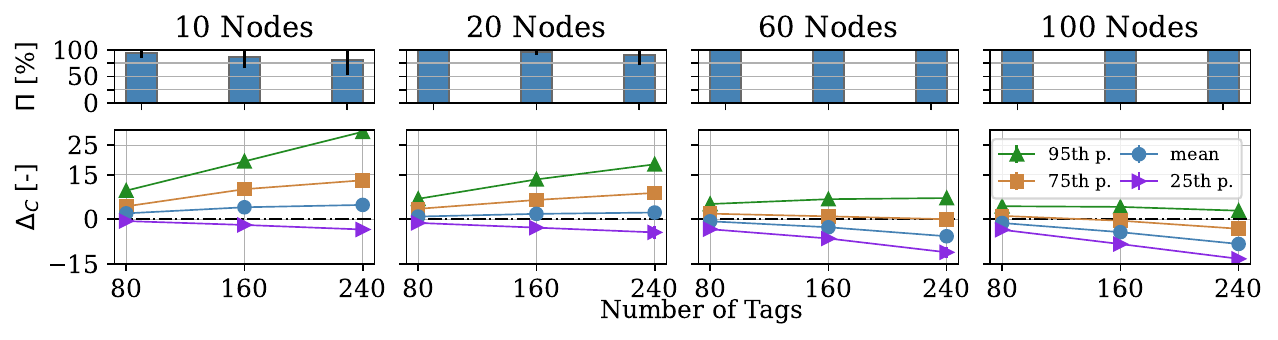}}
  \caption{\textit{Warm-up proven crucial for more stable performance of percentage of correctly computed schedules $\mathbf{S_{c}}$ for larger topologies.} 
     Performance over multiple runs (higher is better). Vertical lines represent the standard error of the mean of the respective metric.}
  \label{fig:warmup} 
\end{figure}

\fakepar{Warmup contributes to higher $\Pi$ values} 
Without warmup, Figure~\ref{subfig:nowarmup} shows how the performance from the percentage of correctly computed schedules $\Pi$ deteriorates (also with increasing std-err) as the topology size increases. 
Including warmup significantly mitigated the variance in $\Pi$ for the larger topologies, as shown in Figure~\ref{subfig:2H}. 
Moreover, it improves Carriers Saved $\boldsymbol{\Delta}_C$ values for the 25th, mean, 75th and 95th percentiles in topologies of up to 60 nodes. However, the average performance of $\boldsymbol{\Delta}_C$ across the multiple runs is similar for 100 node topologies, with only marginal improvements when including warmup. Moreover, including warm-up also reduced the standard error of all metrics (vertical lines), regardless of the topology size.

\subsubsection{Influence of Positional Encoding}
\label{subsec:pos_enc}

We investigate augmenting the input node features to the \ac{GNN} with \acp{PE} to aid the model in breaking symmetries. 
Based on the results from Sec.~\ref{subsec:warm-up}, all models are trained with warmup.
We consider three types of \acp{PE} considering both local and global graph properties. We train four models for each \ac{PE} configuration.

\paragraph{Node Degree PE}
\label{subsec:node_deg}
We include one additional vector in the input node feature matrix that corresponds to the normalized node degree vector. Given the adjacency matrix $\mathbf{A}\in\mathbb{R}^{N\times N}$ of an undirected graph $G=\langle V_a,E \rangle$ with $|V_a|=N$ nodes, where $\mathbf{A}[u, v] = 1$ if $\langle u, v\rangle\in E$ and $A[u, v] = 0$ otherwise, the degree of node $u$ is $\Tilde{d}_u=\sum_{v\in V_a}\textbf{A}[u,v]$~\cite{hamilton2020graphbook}. 
We append the node degree vector $\Tilde{d}=[\Tilde{d}_u/\Tilde{d}_{max}]_{u\in V_a}^{\top}\in\mathbb{R}^N$ as a column to the input node feature matrix $X\in\mathbb{R}^{N\times D}$, where $\Tilde{d}_{max}$ is the degree with highest magnitude. 
Including node degree \ac{PE} results in $D=3+1=4$ input features per node.

\paragraph{Eigenvalues of Graph Laplacian (Eigvals PE)}
We investigate using global properties of the graph as \ac{PE}. 
We define the symmetric normalized graph Laplacian as $\mathbf{L} = I - \mathbf{D}^{-\frac{1}{2}}\mathbf{A}\mathbf{D}^{-\frac{1}{2}}$, where $\mathbf{D}=\diag(\Tilde{d})$ is the diagonal node degree matrix and $I$ is the identity matrix. We perform \ac{EVD} of $\mathbf{L}$ resulting in $\mathbf{L}=\mathbf{V}\mathbf{\Lambda}\mathbf{V}^{-1}$, where $\mathbf{\Lambda}\in\mathbb{R}^{N\times N}$ is a diagonal matrix containing the eigenvalues $\lambda_i\in\mathbb{R}$ of $\mathbf{L}$, and $\mathbf{V}\in\mathbb{R}^{N\times N}$ is a matrix containing the eigenvectors $\mathbf{v}_i\in\mathbb{R}^N$ for $i\in V_a$. 
We first augment the node feature matrix with a vector that contains the eigenvalues of the graph $\Tilde{\Lambda}=[\lambda_i]_{i\in V_a} \in \mathbb{R}^N$. We normalize $\hat{\Lambda}$ using the highest eigenvalue.
Including Eigvals \ac{PE} results in $D=3+1=4$ input features per node.

\paragraph{Stable and Expressive Positional Encodings (SPE PE)}
While eigenvalues provide an indication of magnitude and transformation strength, eigenvectors contain richer geometric information in the directional properties. 
Eigenvectors are not unique, and suffer from sign invariance---i.e., if $\mathbf{v}$ is an eigenvector, so is $-\mathbf{v}$. Geometrically, this means that they are nontrivial solutions for finding the \ac{EVD}: any orthogonal change of basis of $\mathbf{V}$ yields the same Laplacian $\mathbf{L}$~\cite{kwak2004linear}. 

While early work introduces random eigenvector sign flipping during training to account for sign invariance~\cite{Diwedi2023benchmarking, kreuzer2021rethinking}, recent work explores learning the invariances that account for changes in the eigenspace basis $\mathbf{V}$~\cite{lim2022sign,huang2024spe}. The goal is to learn a permutation-invariant transformation of $\Tilde{\Lambda}$ and $\mathbf{V}$ that accounts for their geometrical significance. 
%
We choose the Stable and Expressive \ac{PE} (SPE) method presented by Huang et al.~\cite{huang2024spe} due to its benefits over previous methods~\cite{lim2022sign}. We construct a \ac{PE} matrix $\Gamma\in\mathbb{R}^{N\times Z}$ using the first $Z$ smallest Eigenvalues $\hat{\Lambda}=[\Tilde{\Lambda}_i]_{i\in[0:Z]}\in \mathbb{R}^{Z}$ and Eigenvectors $\hat{\mathbf{V}} = [\mathbf{V}_{[:,i]}]_{i\in[0:Z]}\in\mathbb{R}^{N\times Z}$ as~\cite{huang2024spe}:
\begin{equation}
    \Gamma = \rho \left( \hat{\mathbf{V}}\diag(\phi_1(\hat{\Lambda}))\hat{\mathbf{V}}^{\top}, \dots, \hat{\mathbf{V}}\diag(\phi_m(\hat{\Lambda}))\hat{\mathbf{V}}^{\top} \right) \text{,}
\end{equation}
where $\rho$ is a permutation invariant function and $\{\phi_i\}_{i=1}^m$ are $m$ independent linear transformations. We implement $\rho$ using a Graph Isomorphism Network~\cite{xu2018powerful} and $\phi$ with multi-layer perceptrons, using the same hyperparameters as Huang et al.~\cite{huang2024spe}. However, as we operate on a supervised setting, the choice of $Z$ is determined by the training graph sizes (topologies up to 10 nodes). Hence, we choose the first $Z=9$ eigenvalues larger than $0$ and their eigenvectors.
Including SPE \ac{PE} results in $D=3+Z=12$ input features per node.

\begin{figure} 
    \centering
  \subfloat[2 Heads with warmup and \textbf{SPE \ac{PE}}~\cite{huang2024spe}. \label{subfig:spe-PE}]{%
       \includegraphics[width=0.475\textwidth]{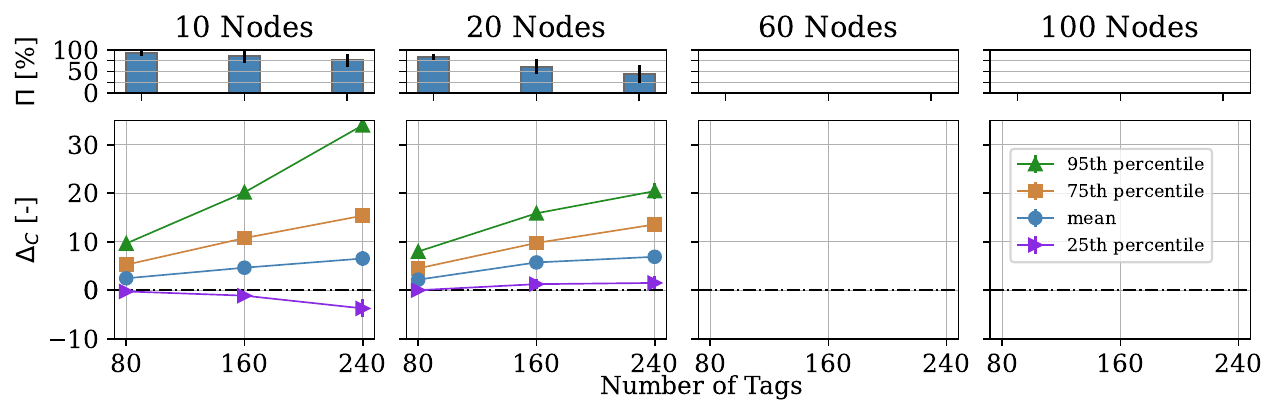}}
    \\
  \subfloat[2 Heads with warmup and \textbf{Eigenvalues \ac{PE}}. \label{subfig:eivals-PE}]{%
        \includegraphics[width=0.475\textwidth]{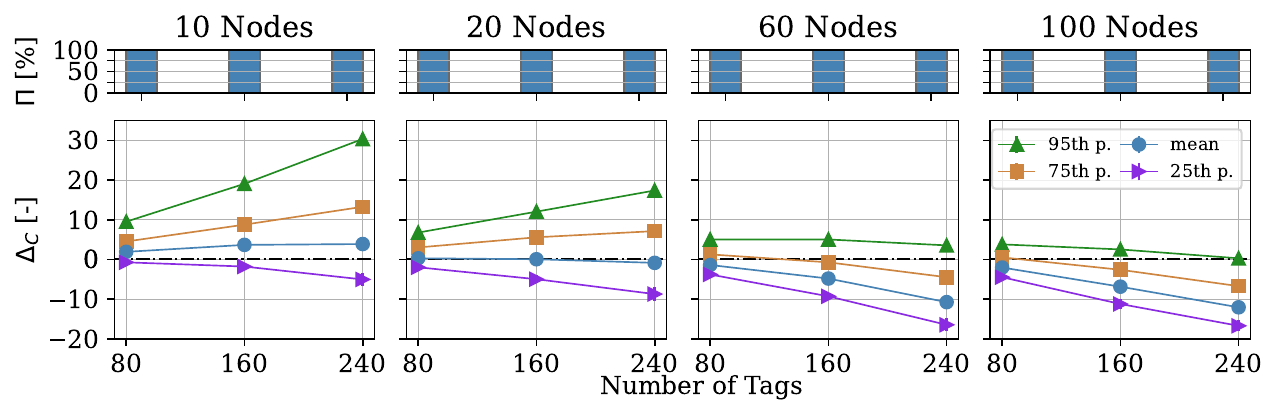}}
    \\
  \subfloat[2 Heads with warmup and \textbf{node degree \ac{PE}}. \label{subfig:degree-PE}]{%
        \includegraphics[width=0.475\textwidth]{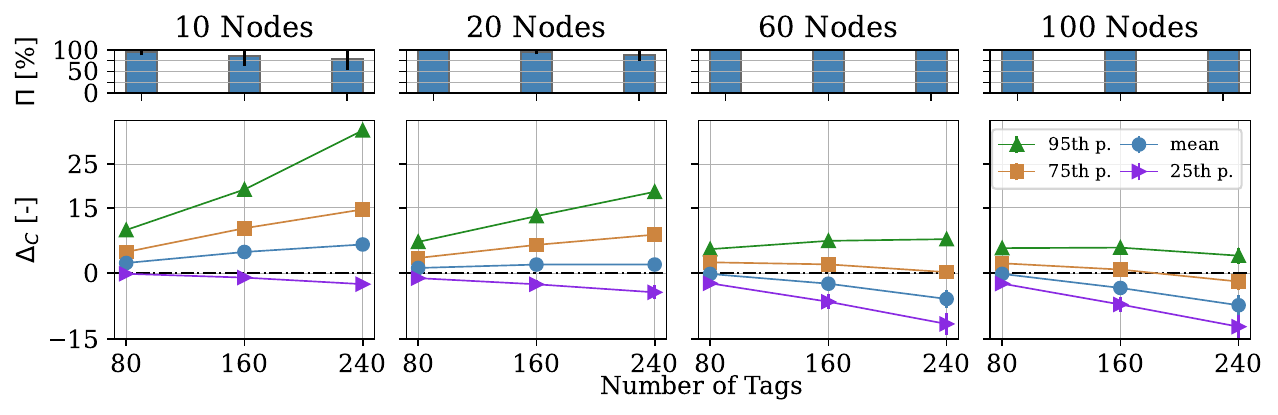}}
  \caption{\textit{Node degree \ac{PE} achieved best trade-off between generalization performance and low computational overhead}.
  Graphs demonstrate influence of different \acp{PE} by showing performance over multiple runs. Vertical lines depict the standard error of the respective metric.}
  \label{fig:pe} 
\end{figure}

\fakepar{Node degree \ac{PE} provides the best trade-off between symmetry-breaking and computational overhead}
Figure~\ref{fig:pe} depicts the model's performance for different \ac{PE} methods. 
While SPE achieved the best carrier saved $\Delta_C$ results in topologies up to 20 nodes (Figure~\ref{subfig:spe-PE}), its $\Pi$ value significantly reduces for an increasing number of sensor tags. Moreover, it is completely unable to compute schedules for topologies of 60 and 100 nodes ($\Pi=0$). Moreover, Figures~\ref{subfig:eivals-PE} and~\ref{subfig:degree-PE} for Eigvals \ac{PE} and node degree \ac{PE} show similar profiles. Notably, node degree \ac{PE} achieves higher 75th and 95th percentile values for both 60 node and 100 node topologies. Additionally, node degree \ac{PE} does not incur in the expensive computation overhead of estimating the \ac{EVD}. E.g., it takes on average 450 ms extra to compute the \ac{EVD} on a multi-core processor for 100 node topologies. Hence, node degree \ac{PE} represents a good trade-off to improve the performance, while avoiding the \ac{EVD} computation overhead.

\subsubsection{Influence of Attention Heads}
\label{subsec:attn_heads}

\begin{figure} 
    \centering
  \subfloat[\textbf{4 Heads} with warmup and node degree \ac{PE}. \label{subfig:4H}]{%
       \includegraphics[width=0.475\textwidth]{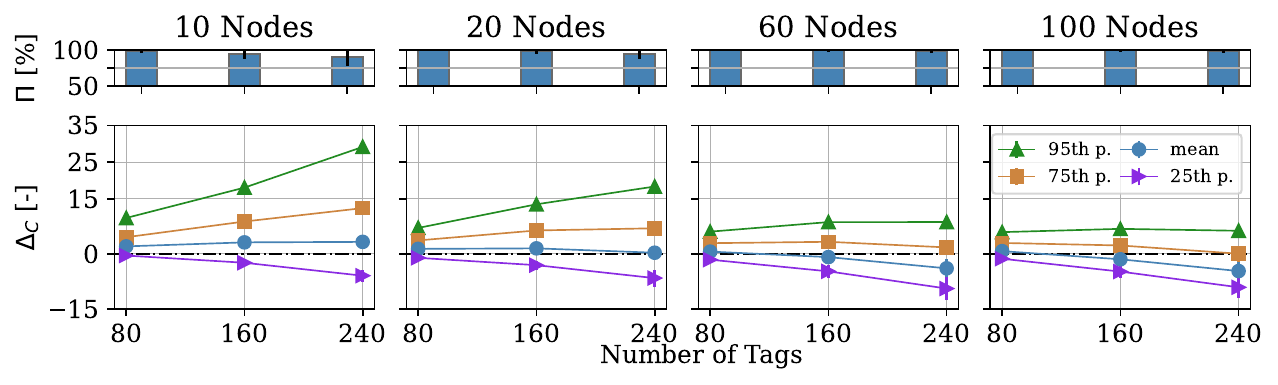}}
    \\
  \subfloat[\textbf{8 Heads} with warmup and node degree \ac{PE}. \label{subfig:8H}]{%
        \includegraphics[width=0.475\textwidth]{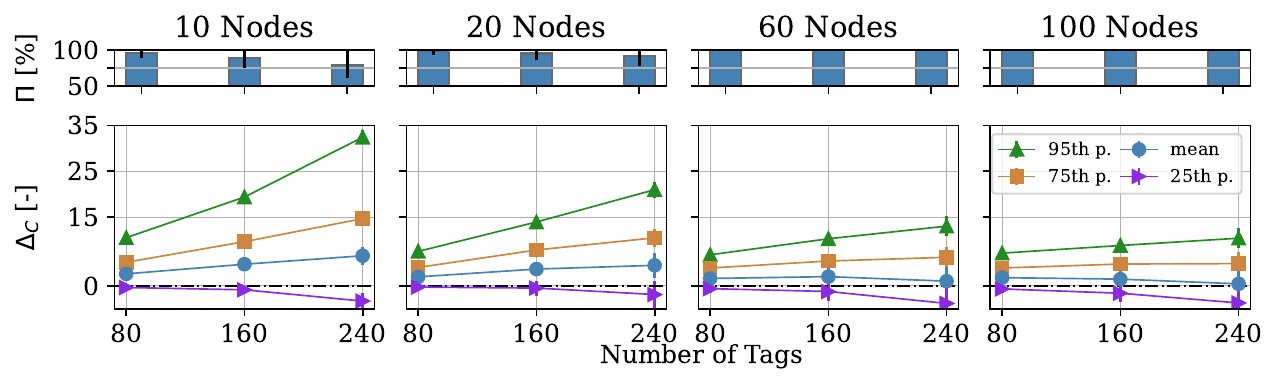}}
    \\
  \subfloat[\textbf{\system}: \textbf{12 Heads}\textbf{ with warmup and node degree \ac{PE}}. \label{subfig:12H}]{%
        \includegraphics[width=0.475\textwidth]{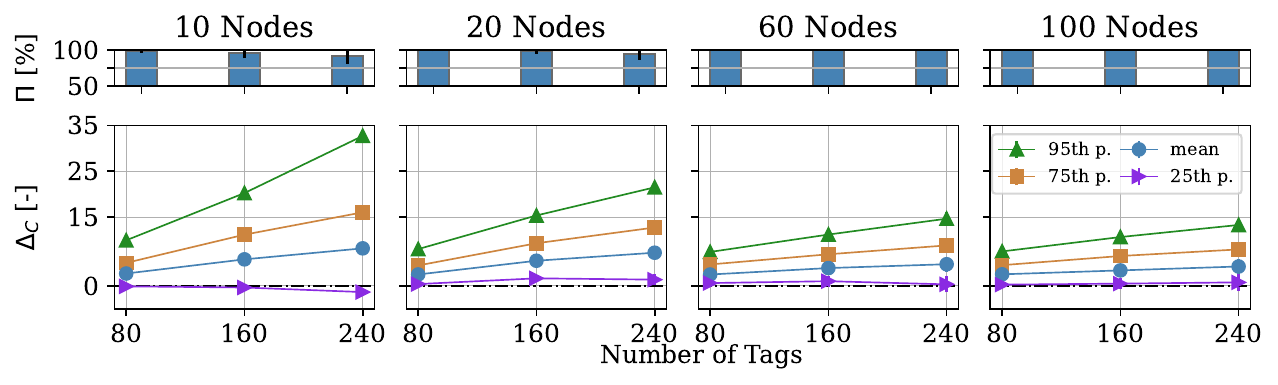}}
    \\
  \subfloat[\textbf{16 Heads} with warmup and node degree \ac{PE}. \label{subfig:16H}]{%
  \includegraphics[width=0.475\textwidth]{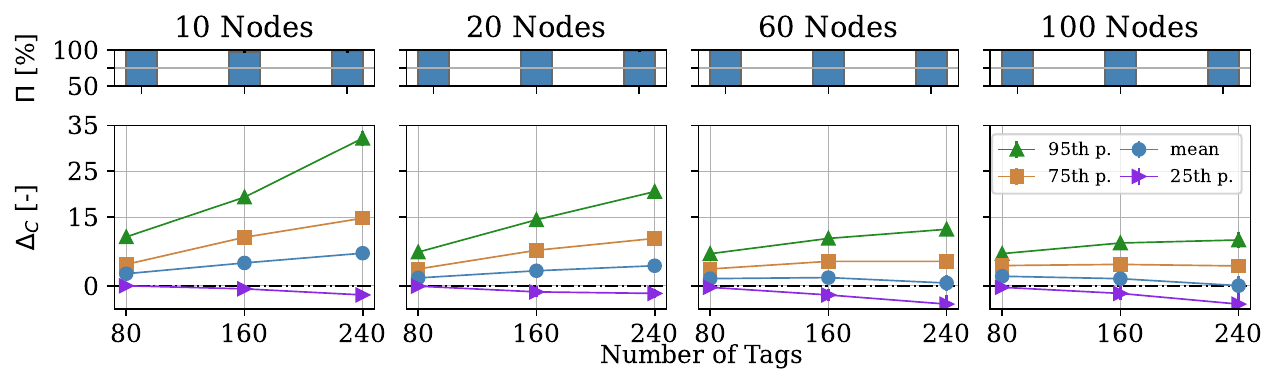}}
  
  \caption{\textit{Robust and consistent generalization achieved with 12 Heads.} 
  Graphs depict the influence of the number of attention heads by showing performance over multiple runs. Vertical lines depict the standard error of the respective metric.}
  \label{fig:attentionheads} 
\end{figure}

We include warmup and node degree \ac{PE} based on the results from the previous sections.
We now evaluate the influence of model complexity by increasing the number of attention heads $M$ in each of the \ac{GNN} layers.  
We train eight models for each number-of-heads value in $M=\{4, 8, 12\}$, and four 16-head models due to their long runtimes (+40 hours per model).
We report average and standard error from performance metrics' statistics.   

\fakeparagraph{12 heads crucial for robust generalization} 
Figure~\ref{fig:attentionheads} shows the influence of increasing the number of attention heads in the model. As observed from Figure~\ref{subfig:4H}-~\ref{subfig:12H}, increasing the attention heads implies an increase in the carriers saved $\Delta_C$ performance for all percentiles. While the models from 8 heads and 12 heads exhibit similar performance, the overall stability of 12 heads is better for both percentage of correctly computed schedules $\Pi$ and for pushing the 25th percentile of $\Delta_C$ above 0. Increasing the attention heads beyond 12 to 16 yields no benefit. On the contrary: Figure~\ref{subfig:16H} shows how the mean and 25th percentile of $\Delta_C$ fall below 0.

\subsection{Final GNN Model} 

Our analysis from Sec.~\ref{subsec:results} results in a \system model of 12 attention heads with node degree \ac{PE} that is trained with warmup. It exhibits strong generalization to larger topologies, and its performance is consistent across independently trained models.
We train \system's model according to Sec.~\ref{subsec:model-trainig} using the dataset described in Sec.~\ref{subsec:training_data}.
Training the model with a mini-batch size of 1024 requires 22 hours on an NVIDIA A100 GPU.


\section{Evaluation}
\label{sec:evaluation}

\begin{figure}
    \centering
    \subfloat[Carriers saved $\Delta_C$ vs TagAlong heuristic. Markers depict mean, box extents delimit 25 and 75 percentiles, and whiskers 1 and 99 percentiles.\label{subfig:robust-vs-deep-CARR}]{
    \includegraphics[width=0.485\textwidth]{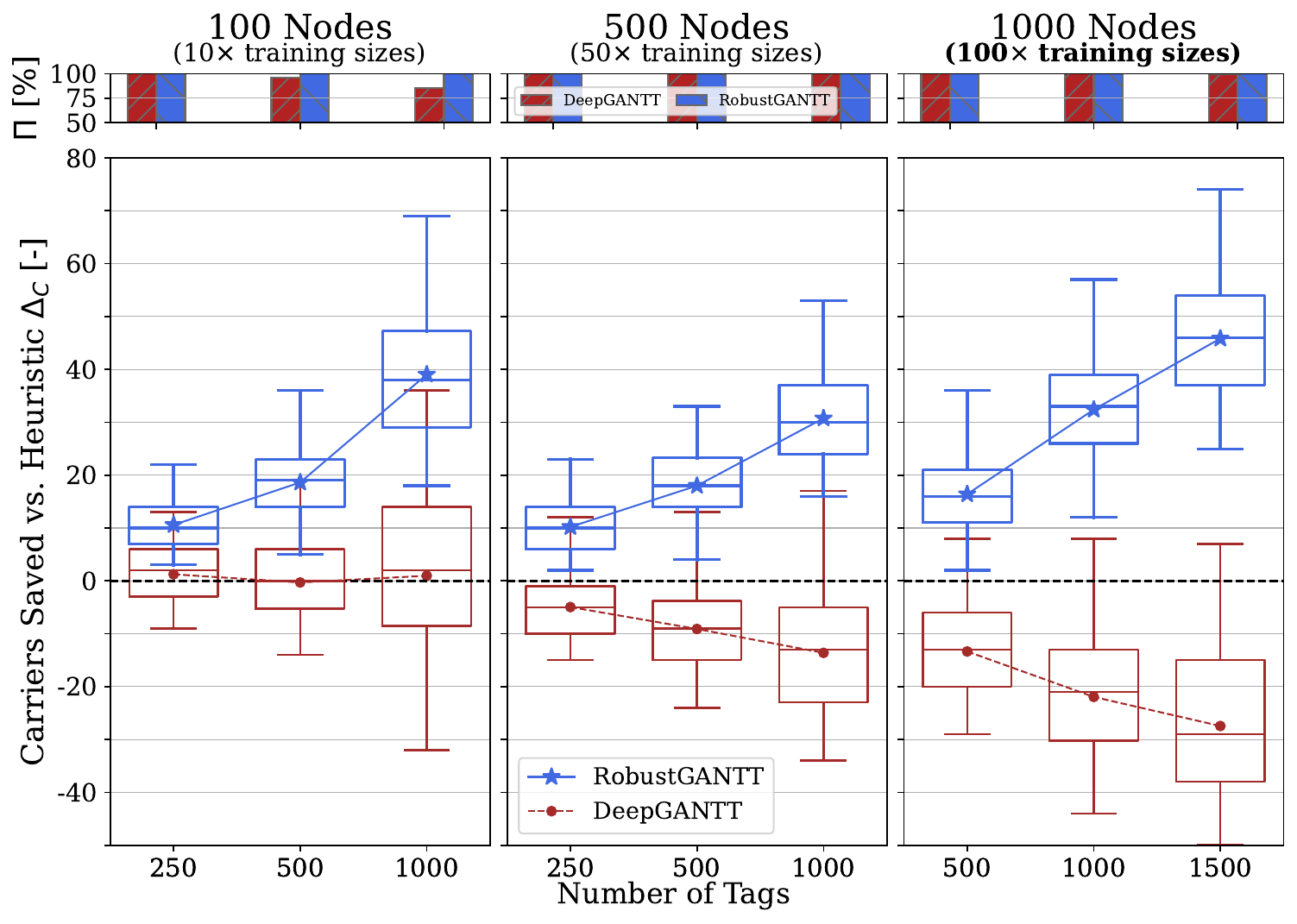}}
    \\
    \subfloat[Timeslots saved $\Delta_L$ vs TagAlong heuristic. Bars depict mean, vertical lines standard deviation.\label{subfig:robust-vs-deep-LEN}]{
    \includegraphics[width=0.485\textwidth]{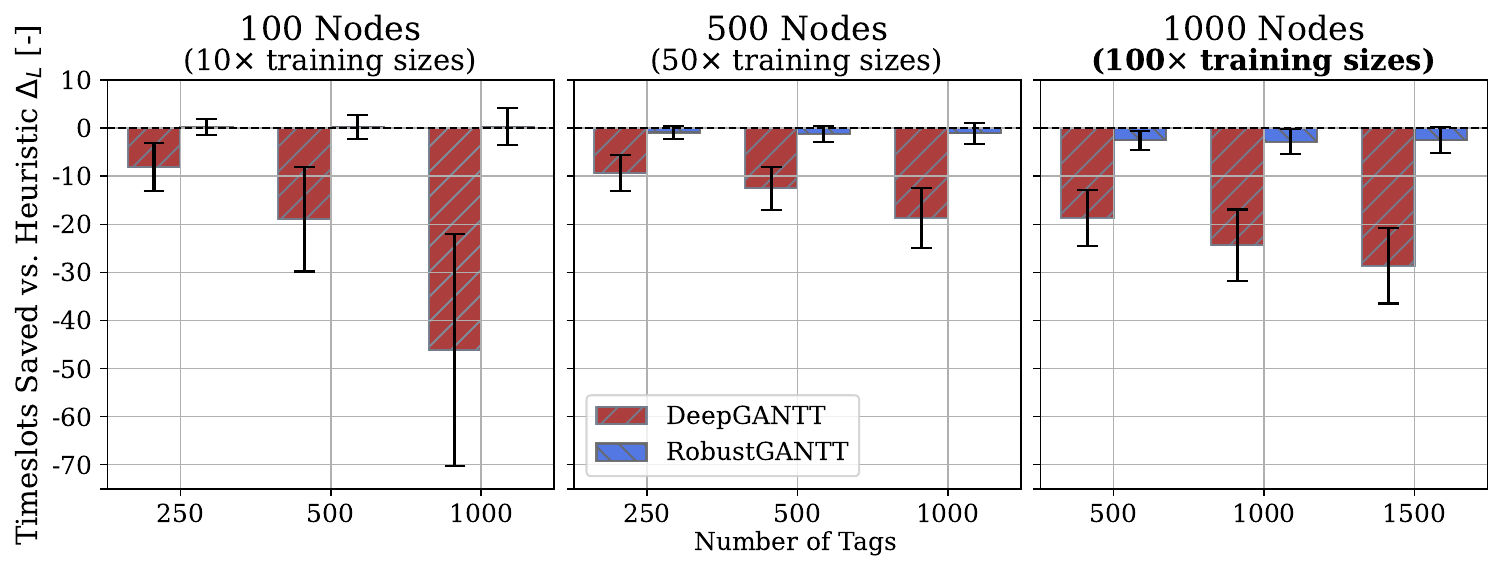}}

    \caption{\system achieves increasing carrier savings with increasing topology sizes compared to the heuristic, while utilizing roughly the same timeslots. Comparison of \oldsys~\cite{perezramirez_DeepGANTT_2023} and our scheduler against the TagAlong heuristic.}
    \label{fig:robust-vs-deep}
\end{figure}

In this section, we compare \system's performance against the \oldsys scheduler in terms of resource savings over the TagAlong heuristic~\cite{PerezPenichet2020afast}. We use both simulated topologies and a real-life \ac{IoT} network. 
The design choice of \acp{GNN} allows our scheduler to generalize to larger, previously unseen network topologies without retraining. Hence, no further \system's \ac{ML} model training is performed for these experiments. We highlight the following \textbf{key findings}: 
\begin{itemize}[leftmargin=*]
    \item \system far surpasses the generalization capabilities of \oldsys. It scales to 1000 node topologies, while increasingly saving resources compared to TagAlong without sacrificing latency (Figure~\ref{fig:robust-vs-deep}).
    \item Both \system and \oldsys achieve resource savings against TagAlong for the real-life \ac{IoT} network. However, our scheduler achieves up to $1.9\times$ more energy savings, up to a $5.7\times$ reduction in the schedule's latency, and up to $2\times$ reduction in 95th percentile runtime compared to \oldsys (Figures~\ref{fig:pi_performance} and \ref{fig:pi_runtimes}).
    \item For the real-life \ac{IoT} network topology, \system achieves an average runtime of 540ms, which allows it to react fast to changing network conditions. 
\end{itemize}

\fakeparagraph{Implementation}
We implement \system as Function as a Service with $\sim2600$ lines of code in a server with an A100 NVIDIA GPU. In general, \system's \ac{ML} model has $\sim295$ million parameters, requiring $\sim1.6$ GB GPU memory in total using single-point precision, which  allows deploying \system in lower-end GPUs.

\subsection{Scalability to 1000-node topologies}

We evaluate \system's generalization to larger topologies, far exceeding \oldsys's capabilities, while still achieving significant resource savings against TagAlong.

\subsubsection{Dataset} We consider 200 problem instances (simulated \ac{IoT} networks with random sensor tag assignments) for $(N, T)$ pairs from the sets $N\in\{100, 500, 1000\}$ nodes and $T\in\{250, 500, 1000, 1500\}$ sensor tags. 

\subsubsection{Performance metrics} 
\label{subsec:delta-L}
Besides $\Pi$ and $\Delta_C$ (see Sec.~\ref{subsec:perf_metrics}), we consider a metric related to the schedule length $L$.

\fakeparagraph{$\boldsymbol{\Delta}_L$---Timeslots Saved} Relates to the latency of querying all sensor tag values in the network. Given a network topology, it compares the length of the schedule produced by TagAlong $L_{ta}$ against the length from the schedule produced by a learning-based scheduler $L_{nn}$ as: $\Delta_L = L_{ta} - L_{nn}$.

\subsubsection{Results}  
Figure~\ref{subfig:robust-vs-deep-CARR} depicts the carriers saved $\Delta_C$ of both \system and \oldsys against the TagAlong heuristic. \system consistently achieves higher savings with both an increase in the number of nodes and number of sensor tags. Notably, even its 1st percentile lies above zero, i.e., for at least $99\%$ of the cases \system achieves savings against TagAlong. Our scheduler achieves on average $12\%$ and up to a $1.4\times$ reduction in the number of carriers compared to TagAlong. Sec.~\ref{subsec:perf_metrics_testbed} demonstrates how $\Delta_C$ directly translates to energy savings. 
The \oldsys scheduler is, however, only marginally better than TagAlong for 100-node topologies, and increasingly worse for larger networks. 
Additionally, \oldsys's correctly computed schedules $\Pi$ decreases for 100 nodes, while \system's values are consistently $\Pi=100\%$.

\system computes schedules requiring roughly the same number of timeslots as TagAlong ($\Delta_L\approx 0$) as shown in Figure~\ref{subfig:robust-vs-deep-LEN}. Hence, our scheduler achieves significant savings in energy and spectrum without a significant reduction in the latency to query all sensor tags. Across all topologies considered, \system requires on average 1.12 additional timeslots compared to TagAlong. In contrast, \oldsys requires on average 20 additional timeslots, and achieves no resource savings for such large topologies. Moreover, Figure~\ref{subfig:robust-vs-deep-LEN} shows how \oldsys requires  increasingly more timeslots than our scheduler.

\subsection{Performance for a Real IoT Network}

\begin{figure}
    \centering
    \includegraphics[width=0.485\textwidth]{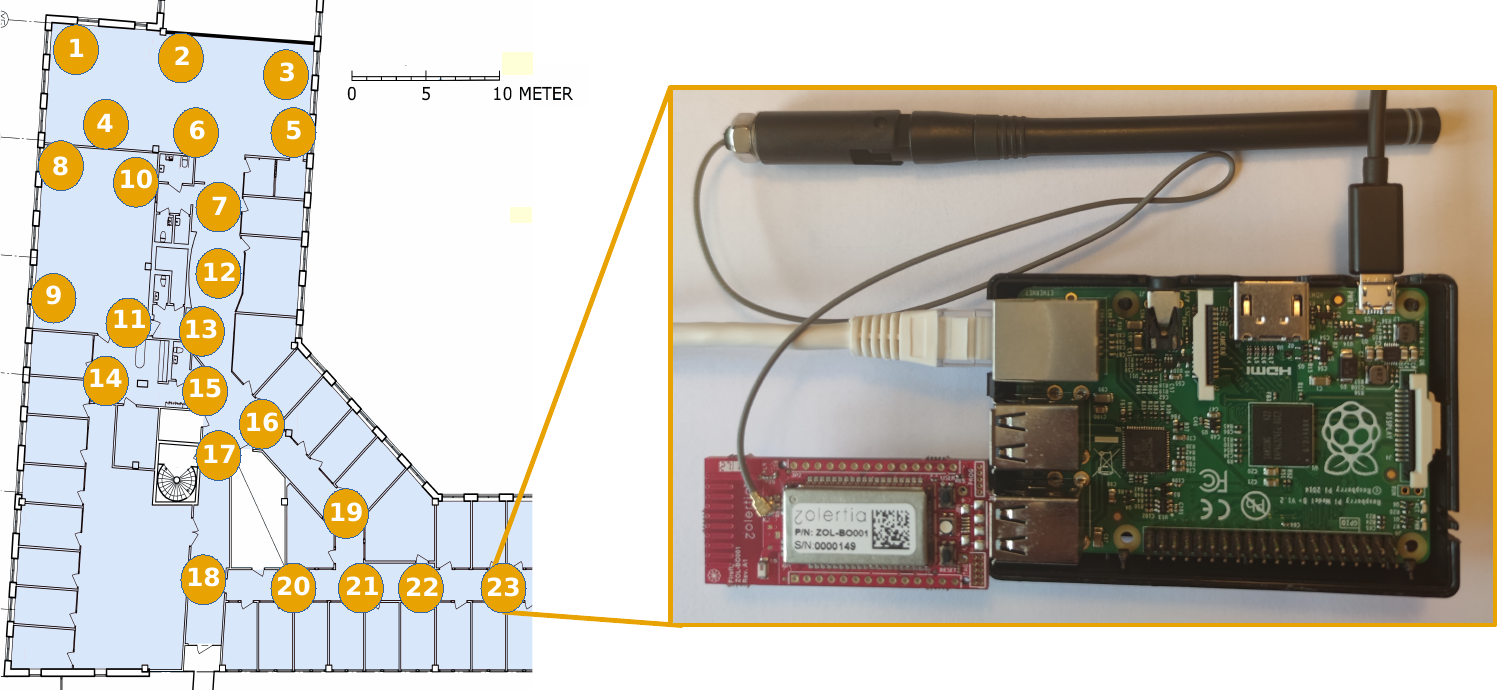}
    \caption{Floor layout and device configuration of our testbed. We use a real \ac{IoT} network consisting of Zolertia Fireflies, and collect network metrics over a period of four days.}
    \label{fig:real-iot}
\end{figure}

We now evaluate \system's ability to compute schedules for a real-life \ac{IoT} network. 

\subsubsection{Testbed} Our experimental setup utilizes an indoor \ac{IoT} testbed composed of 23 Zolertia Firefly devices running Contiki-NG~\cite{oikonomou22contiking} (see Figure~\ref{fig:real-iot}). These devices employ the RPL routing protocol~\cite{winter2012rpl} and communicate via IPv6 over IEEE 802.15.4 \ac{TSCH}~\cite{duquennoy17tsch}. Link connectivity data between \ac{IoT} nodes was gathered at 30-minute intervals across a four-day span, assuming a link exists between node pairs when the signal strength reaches at least \SI{-75}dBm, suitable for carrier provisioning. Additionally, we enhanced each network topology by assigning simulated tags randomly to achieve various densities, defined as $N/T = 23/T$, for $T \in \{46, 115, 230, 460\}$, with each density configuration tested 100 times.

\subsubsection{Performance metrics}
\label{subsec:perf_metrics_testbed}
Besides $\Pi$, $\Delta_C$ (see Sec.~\ref{subsec:perf_metrics}), and $\Delta_L$ (see Sec.~\ref{subsec:delta-L}) we explicitly evaluate energy consumption. Moreover, percentages for $\Delta_C$ and $\Delta_L$ imply normalization with respect to the heuristic values, e.g., $\Delta_{C\%} = \Delta_C / C_{ta}$.

\fakepar{$\boldsymbol{\Delta}_{\mathbf{E\%}}$---Energy Saved} We consider the average energy required for querying the tag's sensor values $\Tilde{E}$. It corresponds to the total energy required to interrogate all sensor tags $E_{tot}$ divided by the number of tags $T$ in the network~\cite{perezramirez_DeepGANTT_2023}: 
\vspace{-0.1cm}
\begin{eqnarray}
    \Tilde{E} \!\!\!&\!\!\!=\!\!\!&\!\!\! \frac{E_{tot}}{T} 
    \!=\! P_{tx}t_{tx}\!+\!P_{rx}\left(\frac{C}{T}t_{req}\!+\!t_{rx}\right)\!+\!P_{tx}\left(t_{req}\!+\!\frac{C}{T}t_{cg}\right) \!\text{,}
    \label{eq:avg-energy-pertag}
\end{eqnarray}
where both $P_{rx}$ and $P_{rx}$ correspond to the radio power at transmit and receive mode, respectively. 
$t_{rx}$, $t_{tx}$, $t_{req}$, and $t_{cg}$ are defined as in Figure~\ref{fig:tag-to-host-comm}. 
Calculating a percentage of energy saved against the TagAlong scheduler corresponds to $\Delta_{E\%}=(\Tilde{E}_{nn}-\Tilde{E}_{ta})/\Tilde{E}_{ta}$. 
Given a schedule, all values in Eq.~\ref{eq:avg-energy-pertag} except $C$ are constant for calculating both $\Tilde{E}_{ta}$ and $\Tilde{E}_{nn}$. Hence, lower values of $C$ directly translates to energy savings. 

We adopt $P_{rx}=72mW$, $P_{tx}=102mW$ based on the Firefly's reference values. Moreover, we assume $t_{req}=t_{tx}=128\mu s$, $t_{rx}=256\mu s$, and $t_{cg}=15.75 ms$~\cite{PerezPenichet2020afast, perezramirez_DeepGANTT_2023}. 

\begin{figure}
    \centering
    \subfloat[Carriers saved.\label{subfig:pi_diff_carr}]{%
       \includegraphics[width=0.235\textwidth]{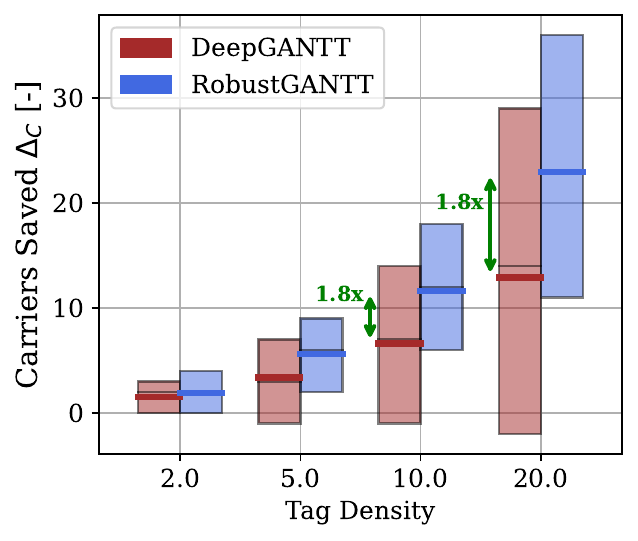}}
    \hfill
    \subfloat[Timeslots saved. 
    \label{subfig:pi_timeslots_saved}]{%
        \includegraphics[width=0.235\textwidth]{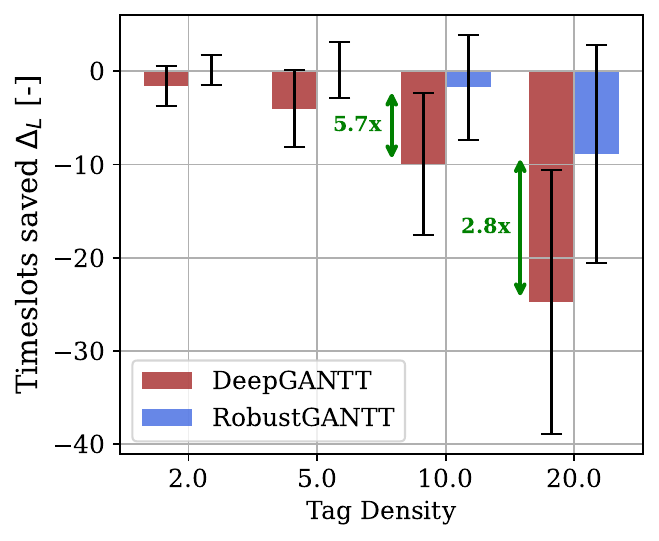}}
    \\
    \subfloat[Percentage of carriers saved. 
    \label{subfig:pi_perc_diff_carr}]{%
        \includegraphics[width=0.235\textwidth]{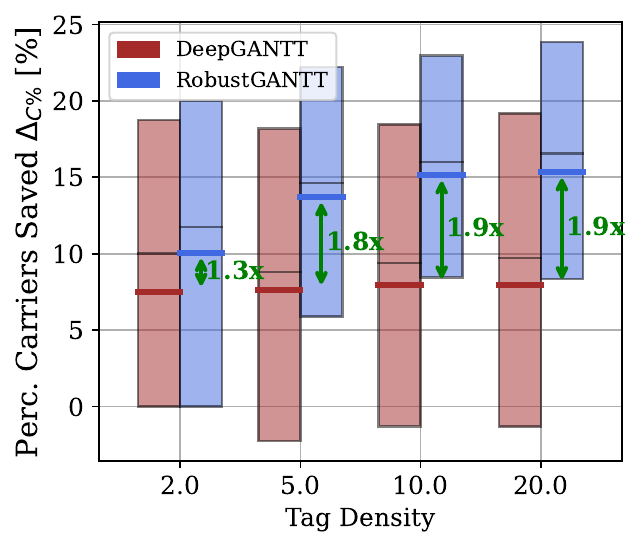}}
    \hfill
    \subfloat[Energy saved. 
    \label{subfig:pi_perc_energy}]{%
        \includegraphics[width=0.235\textwidth]{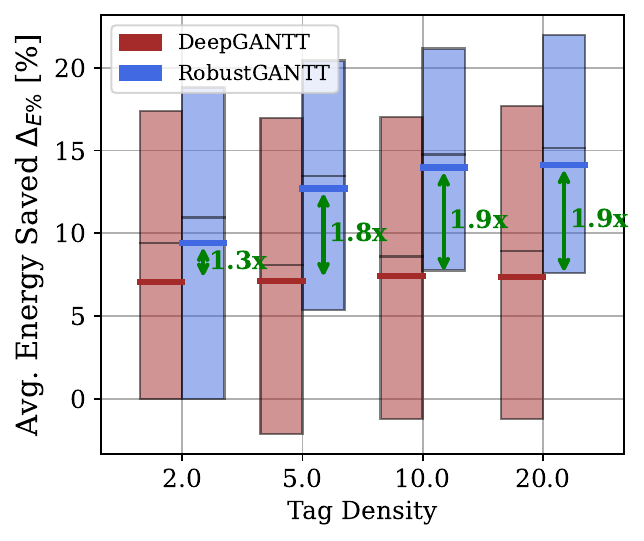}}
        
    \caption{
    \system achieves more resource savings than \oldsys against TagAlong for a real IoT network, while requiring roughly as many timeslots.
    Performance comparison for a real-life \ac{IoT} network topology of both \system and \oldsys against the TagAlong heuristic. Boxplots depict the mean and interquartile range. Bar plot depict the mean and std-dev.} 
    \label{fig:pi_performance}
\end{figure}

\begin{figure}
    \centering
    \includegraphics[width=0.48\textwidth]{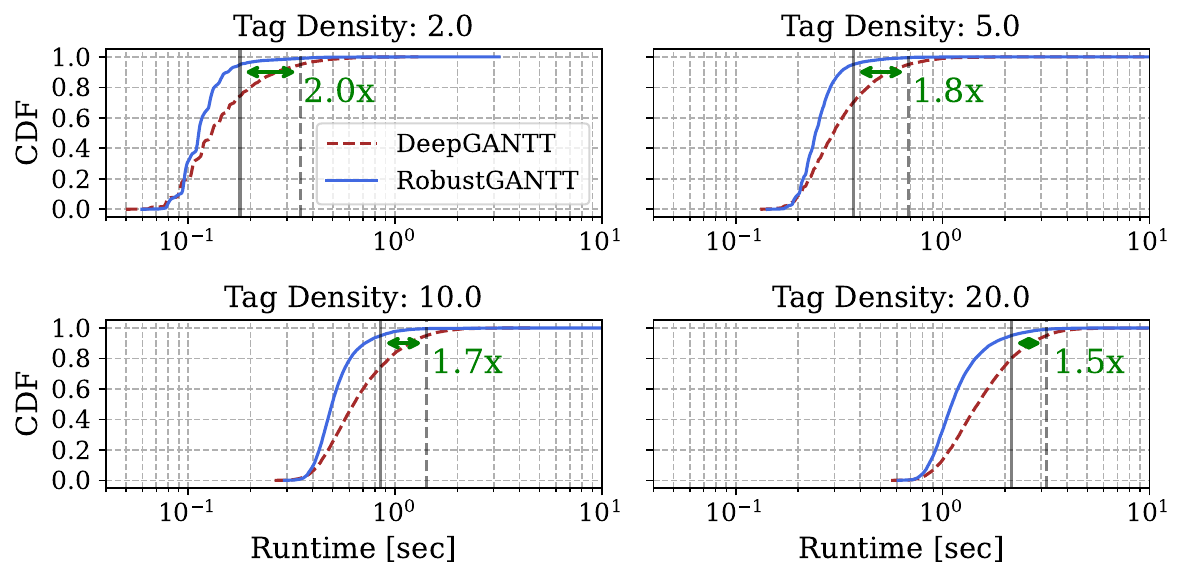}
    \caption{\system exhibits up to $2\times$ reduction in 95th percentile runtime values compared to \oldsys. Runtimes for different tag densities $N/T$.}
    \label{fig:pi_runtimes}
\end{figure}

\subsubsection{Results}
For the real-life IoT network, \system achieves on average $14\%$ and up to $53\%$ energy savings $\Delta_{E\%}$ (i.e., up to $2\times$ less energy) compared to TagAlong. Even for the highest tag densities $N/T$ considered, our scheduler achieves $44\%$ energy savings. Such savings represent up to $1.9\times$ the savings achieved by \oldsys, as shown in Figure~\ref{subfig:pi_perc_energy}. Figures~\ref{subfig:pi_diff_carr} and~\ref{subfig:pi_perc_energy} demonstrate the equivalence between $\Delta_{C\%}$ and $\Delta_{E\%}$: a reduction in the number of carriers directly translates to energy savings.

In terms of latency to query all sensor values, Figure~\ref{subfig:pi_timeslots_saved} shows how \oldsys always requires on average more timeslots than TagAlong. In contrast, our scheduler requires on average as many timeslots as TagAlong for tag densities 2.0 and 5.0, and 10. However, it requires on average 8.8 more timeslots for tag density 20.
Figure~\ref{fig:pi_runtimes} shows the runtime distributions of both schedulers across tag densities. Their profiles are those of heavy-tailed distributions. 
Both schedulers require roughly the same average runtimes across tag densities. In particular, \system's average runtimes are 120 ms, 260 ms, 540 ms, and 1.2 sec for the respective tag densities 2, 5, 10, and 20. However, Figure~\ref{fig:pi_runtimes} demonstrates how \system reduces the runtime's 95th percentile up to a factor of $2\times$ compared to \oldsys.

While the real-life network's size is within \oldsys's proven generalization capabilities~\cite{perezramirez_DeepGANTT_2023}, we demonstrate that our scheduler requires on average up to $1.9\times$ less carriers (energy savings), up to $5.7\times$ less timeslots (reduction in latency to query all sensor tags), and up to a $2\times$ reduction in 95th percentile runtime to compute the schedule.

\section{Discussion}
\label{sec:discussion}
\system is a scheduler that far surpasses the generalization capabilities of existing learning-based systems. Our system can not only processes much larger \ac{IoT} network topologies than previously possible, but also delivers more resource-efficient schedules.   

\fakeparagraph{Large-scale IoT networks} Our system is designed to reduce energy consumption in IoT networks. This is of paramount importance not only for sustainability reasons, but also because such networks are typically energy constrained. Moreover, ensuring energy savings without increasing querying latency is highly relevant, specially for dense network deployments, since it reduces spectrum utilization.  

\fakeparagraph{Serving IoT networks in parallel}
The NP-Hard nature of generating resource-efficient schedules requires deploying \system at the Edge/Cloud, which is also true for other schedulers~\cite{PerezPenichet2020afast, perezramirez_DeepGANTT_2023}. However, one does not require deploying a \system scheduler for every \ac{IoT} network. Rather, one \system instance can process requests from multiple \ac{IoT} networks either in sequence, or by batching those requests and computing their schedules in parallel. 
However, the number of requests processed in parallel is limited by the total the amount of GPU memory available.   

\fakeparagraph{Latency to query all sensor values} \system's schedules require roughly the same number of timeslots as those produced by TagAlong (see Figures~\ref{subfig:robust-vs-deep-LEN} and ~\ref{subfig:pi_timeslots_saved}). This implies that our system does not sacrifice querying latency to achieve its significant energy savings. However, there are cases in which TagAlong schedules are shorter than those from \system. We attribute this to the optimization objective (Eq.~\ref{eq:optobj}), which prioritizes reducing the number of carriers, since we are most interested in energy savings.
Moreover, we do not envision backscatter sensor tags to assist in time-critical settings, but rather in energy-efficient sensing and monitoring. 

\fakeparagraph{Dynamic Environments} Our system exhibits average runtimes of hundreds of milliseconds, allowing it to react fast to connectivity changes in the \ac{IoT} devices. Similarly, adding or removing IoT nodes would trigger a new request to compute a schedule. 
However, detecting the addition or removal of sensor tags to the \ac{IoT} nodes is a general problem for the type of backscatter networks considered, and lies outside our scope.  

\section{Conclusion}
\label{sec:conclusion}
We present \system, a novel system that leverages the latest advancements in \acp{GNN} and \ac{ML} to schedule communications in an \ac{IoT} network augmented with  backscatter sensor tags. We exploit our system design choice of using \ac{GNN} to train our scheduler using optimal schedules from small networks of up to 10 nodes, and demonstrate that \system can seamlessly generalize without re-training to networks of up to 1000 nodes. Our scheduler surpasses the generalization capabilities of current learning-based systems, while achieving significant savings in energy usage, spectrum utilization, and compute runtime.
\system facilitates the large-scale integration of \ac{IoT} networks with sensor tags, and significantly reduces their operational expenses by efficiently utilizing their resources.  

\begin{acks}
This work was financially supported by the Swedish Foundation for Strategic Research (SSF). We acknowledge the usage of High-Performance Computing resources under the EuroHPC JU project No. EHPC-DEV-2023D08-049.
We also thank M.Sc. Peder Hårderup for initial concept prototyping of the node degree \ac{PE} during his M.Sc. thesis at RISE.
\end{acks}

\bibliographystyle{ACM-Reference-Format}
\bibliography{refs}


\begin{thebibliography}{63}


\ifx \showCODEN    \undefined \def \showCODEN     #1{\unskip}     \fi
\ifx \showDOI      \undefined \def \showDOI       #1{#1}\fi
\ifx \showISBNx    \undefined \def \showISBNx     #1{\unskip}     \fi
\ifx \showISBNxiii \undefined \def \showISBNxiii  #1{\unskip}     \fi
\ifx \showISSN     \undefined \def \showISSN      #1{\unskip}     \fi
\ifx \showLCCN     \undefined \def \showLCCN      #1{\unskip}     \fi
\ifx \shownote     \undefined \def \shownote      #1{#1}          \fi
\ifx \showarticletitle \undefined \def \showarticletitle #1{#1}   \fi
\ifx \showURL      \undefined \def \showURL       {\relax}        \fi
\providecommand\bibfield[2]{#2}
\providecommand\bibinfo[2]{#2}
\providecommand\natexlab[1]{#1}
\providecommand\showeprint[2][]{arXiv:#2}

\bibitem[Ahmad et~al\mbox{.}(2021)]%
        {ahmad2021enabling}
\bibfield{author}{\bibinfo{person}{Abeer Ahmad}, \bibinfo{person}{Xiao Sha}, \bibinfo{person}{Milutin Stana{\'c}evi{\'c}}, \bibinfo{person}{Akshay Athalye}, \bibinfo{person}{Petar~M Djuri{\'c}}, {and} \bibinfo{person}{Samir~R Das}.} \bibinfo{year}{2021}\natexlab{}.
\newblock \showarticletitle{Enabling passive backscatter tag localization without active receivers}. In \bibinfo{booktitle}{\emph{Proc. of the 19th ACM Conference on Embedded Networked Sensor Systems (SenSys)}}. \bibinfo{pages}{178--191}.
\newblock


\bibitem[Belkin and Niyogi(2003)]%
        {belkin2003laplacian}
\bibfield{author}{\bibinfo{person}{Mikhail Belkin} {and} \bibinfo{person}{Partha Niyogi}.} \bibinfo{year}{2003}\natexlab{}.
\newblock \showarticletitle{Laplacian eigenmaps for dimensionality reduction and data representation}.
\newblock \bibinfo{journal}{\emph{Neural computation}} \bibinfo{volume}{15}, \bibinfo{number}{6} (\bibinfo{year}{2003}), \bibinfo{pages}{1373--1396}.
\newblock


\bibitem[Bengio et~al\mbox{.}(2021)]%
        {Bengio2021mlforcop}
\bibfield{author}{\bibinfo{person}{Yoshua Bengio}, \bibinfo{person}{Andrea Lodi}, {and} \bibinfo{person}{Antoine Prouvost}.} \bibinfo{year}{2021}\natexlab{}.
\newblock \showarticletitle{{Machine learning for combinatorial optimization: A methodological tour d'horizon}}.
\newblock \bibinfo{journal}{\emph{Eur. J. Oper. Res.}} \bibinfo{volume}{290}, \bibinfo{number}{2} (\bibinfo{date}{apr} \bibinfo{year}{2021}), \bibinfo{pages}{405--421}.
\newblock
\showISSN{03772217}
\urldef\tempurl%
\url{https://doi.org/10.1016/j.ejor.2020.07.063}
\showDOI{\tempurl}
\showeprint[arxiv]{1811.06128}


\bibitem[{Bluetooth SIG}(2021)]%
        {bluetooth_2021}
\bibfield{author}{\bibinfo{person}{{Bluetooth SIG}}.} \bibinfo{year}{2021}\natexlab{}.
\newblock \bibinfo{booktitle}{\emph{Bluetooth {Core} {Specification} 5.3}}.
\newblock


\bibitem[Dai et~al\mbox{.}(2017)]%
        {Dai2017learningcops}
\bibfield{author}{\bibinfo{person}{Hanjun Dai}, \bibinfo{person}{Elias~B. Khalil}, \bibinfo{person}{Yuyu Zhang}, \bibinfo{person}{Bistra Dilkina}, {and} \bibinfo{person}{Le Song}.} \bibinfo{year}{2017}\natexlab{}.
\newblock \showarticletitle{{Learning Combinatorial Optimization Algorithms over Graphs}}. In \bibinfo{booktitle}{\emph{Proc. Advances Neural Inf. Process. Syst. (NIPS)}}, Vol.~\bibinfo{volume}{2017-Decem}. \bibinfo{publisher}{Neural information processing systems foundation}, \bibinfo{pages}{6349--6359}.
\newblock
\showeprint[arxiv]{1704.01665}


\bibitem[Devlin et~al\mbox{.}(2018)]%
        {devlin2018bert}
\bibfield{author}{\bibinfo{person}{Jacob Devlin}, \bibinfo{person}{Ming-Wei Chang}, \bibinfo{person}{Kenton Lee}, {and} \bibinfo{person}{Kristina Toutanova}.} \bibinfo{year}{2018}\natexlab{}.
\newblock \showarticletitle{Bert: Pre-training of deep bidirectional transformers for language understanding}.
\newblock \bibinfo{journal}{\emph{arXiv preprint arXiv:1810.04805}} (\bibinfo{year}{2018}).
\newblock


\bibitem[Duquennoy et~al\mbox{.}(2015)]%
        {duquennoy2015orchestra}
\bibfield{author}{\bibinfo{person}{Simon Duquennoy}, \bibinfo{person}{Beshr Al~Nahas}, \bibinfo{person}{Olaf Landsiedel}, {and} \bibinfo{person}{Thomas Watteyne}.} \bibinfo{year}{2015}\natexlab{}.
\newblock \showarticletitle{Orchestra: Robust mesh networks through autonomously scheduled TSCH}. In \bibinfo{booktitle}{\emph{Proceedings of the 13th ACM conference on embedded networked sensor systems}}. \bibinfo{pages}{337--350}.
\newblock


\bibitem[Duquennoy et~al\mbox{.}(2017)]%
        {duquennoy17tsch}
\bibfield{author}{\bibinfo{person}{Simon Duquennoy}, \bibinfo{person}{Atis Elsts}, \bibinfo{person}{Beshr~Al Nahas}, {and} \bibinfo{person}{George Oikonomou}.} \bibinfo{year}{2017}\natexlab{}.
\newblock \showarticletitle{TSCH and 6TiSCH for Contiki: Challenges, Design and Evaluation}. In \bibinfo{booktitle}{\emph{2017 13th International Conference on Distributed Computing in Sensor Systems (DCOSS)}}. \bibinfo{pages}{11--18}.
\newblock
\urldef\tempurl%
\url{https://doi.org/10.1109/DCOSS.2017.29}
\showDOI{\tempurl}


\bibitem[Dwivedi et~al\mbox{.}(2023)]%
        {Diwedi2023benchmarking}
\bibfield{author}{\bibinfo{person}{Vijay~Prakash Dwivedi}, \bibinfo{person}{Chaitanya~K. Joshi}, \bibinfo{person}{Anh~Tuan Luu}, \bibinfo{person}{Thomas Laurent}, \bibinfo{person}{Yoshua Bengio}, {and} \bibinfo{person}{Xavier Bresson}.} \bibinfo{year}{2023}\natexlab{}.
\newblock \showarticletitle{Benchmarking Graph Neural Networks}.
\newblock \bibinfo{journal}{\emph{Journal of Machine Learning Research}} \bibinfo{volume}{24}, \bibinfo{number}{43} (\bibinfo{year}{2023}), \bibinfo{pages}{1--48}.
\newblock
\urldef\tempurl%
\url{http://jmlr.org/papers/v24/22-0567.html}
\showURL{%
\tempurl}


\bibitem[Ensworth and Reynolds(2015)]%
        {ensworth_every_2015}
\bibfield{author}{\bibinfo{person}{Joshua Ensworth} {and} \bibinfo{person}{Matthew~S. Reynolds}.} \bibinfo{year}{2015}\natexlab{}.
\newblock \showarticletitle{Every smart phone is a backscatter reader: {Modulated} backscatter compatibility with {Bluetooth} 4.0 {Low} {Energy} ({BLE}) devices}. In \bibinfo{booktitle}{\emph{Proc. Ann. Conf. RFID}}. \bibinfo{publisher}{IEEE}.
\newblock


\bibitem[Ferrari et~al\mbox{.}(2011)]%
        {ferrari2011glossy}
\bibfield{author}{\bibinfo{person}{Federico Ferrari}, \bibinfo{person}{Marco Zimmerling}, \bibinfo{person}{Lothar Thiele}, {and} \bibinfo{person}{Olga Saukh}.} \bibinfo{year}{2011}\natexlab{}.
\newblock \showarticletitle{Efficient network flooding and time synchronization with Glossy}. In \bibinfo{booktitle}{\emph{Proc. 10th ACM/IEEE Int. Conf. Information Processing in Sensor Networks}}. \bibinfo{pages}{73--84}.
\newblock


\bibitem[Foundation(2024)]%
        {pytorchReprod2024}
\bibfield{author}{\bibinfo{person}{The~PyTorch Foundation}.} \bibinfo{year}{2024}\natexlab{}.
\newblock \bibinfo{title}{PyTorch Reproducibility}.
\newblock
\newblock
\urldef\tempurl%
\url{https://pytorch.org/docs/stable/notes/randomness.html}
\showURL{%
\tempurl}


\bibitem[Geissdoerfer and Zimmerling(2021)]%
        {geissdoerfer_bootstrapping_2021}
\bibfield{author}{\bibinfo{person}{Kai Geissdoerfer} {and} \bibinfo{person}{Marco Zimmerling}.} \bibinfo{year}{2021}\natexlab{}.
\newblock \showarticletitle{Bootstrapping {Battery}-free {Wireless} {Networks}: {Efficient} {Neighbor} {Discovery} and {Synchronization} in the {Face} of {Intermittency}}. In \bibinfo{booktitle}{\emph{(NSDI'21)}}. \bibinfo{pages}{439--455}.
\newblock
\showISBNx{978-1-939133-21-2}


\bibitem[Gilmer et~al\mbox{.}(2017)]%
        {Gilmer2017mpnn}
\bibfield{author}{\bibinfo{person}{Justin Gilmer}, \bibinfo{person}{Samuel~S. Schoenholz}, \bibinfo{person}{Patrick~F. Riley}, \bibinfo{person}{Oriol Vinyals}, {and} \bibinfo{person}{George~E. Dahl}.} \bibinfo{year}{2017}\natexlab{}.
\newblock \showarticletitle{{Neural Message Passing for Quantum Chemistry}}.
\newblock \bibinfo{journal}{\emph{Proc. 34th Int. Conf. Mach. Learn. (ICML)}}  \bibinfo{volume}{3} (\bibinfo{date}{apr} \bibinfo{year}{2017}), \bibinfo{pages}{2053--2070}.
\newblock
\showeprint[arxiv]{1704.01212}


\bibitem[Guo et~al\mbox{.}(2020)]%
        {guo2020aloba}
\bibfield{author}{\bibinfo{person}{Xiuzhen Guo}, \bibinfo{person}{Longfei Shangguan}, \bibinfo{person}{Yuan He}, \bibinfo{person}{Jia Zhang}, \bibinfo{person}{Haotian Jiang}, \bibinfo{person}{Awais~Ahmad Siddiqi}, {and} \bibinfo{person}{Yunhao Liu}.} \bibinfo{year}{2020}\natexlab{}.
\newblock \showarticletitle{Aloba: Rethinking ON-OFF keying modulation for ambient LoRa backscatter}. In \bibinfo{booktitle}{\emph{Proceedings of the 18th conference on embedded networked sensor systems}}. \bibinfo{pages}{192--204}.
\newblock


\bibitem[Hamilton(2020a)]%
        {hamilton2020graph}
\bibfield{author}{\bibinfo{person}{William~L Hamilton}.} \bibinfo{year}{2020}\natexlab{a}.
\newblock \bibinfo{booktitle}{\emph{Graph representation learning}}. Vol.~\bibinfo{volume}{14}.
\newblock \bibinfo{publisher}{Morgan \& Claypool Publishers}.
\newblock


\bibitem[Hamilton(2020b)]%
        {hamilton2020graphbook}
\bibfield{author}{\bibinfo{person}{William~L Hamilton}.} \bibinfo{year}{2020}\natexlab{b}.
\newblock \bibinfo{booktitle}{\emph{Graph representation learning}}.
\newblock \bibinfo{publisher}{Morgan \& Claypool Publishers}.
\newblock


\bibitem[Hamilton et~al\mbox{.}(2017)]%
        {Hamilton2017inductive}
\bibfield{author}{\bibinfo{person}{William~L. Hamilton}, \bibinfo{person}{Rex Ying}, {and} \bibinfo{person}{Jure Leskovec}.} \bibinfo{year}{2017}\natexlab{}.
\newblock \showarticletitle{{Inductive Representation Learning on Large Graphs}}. In \bibinfo{booktitle}{\emph{Proc. Advances Neural Inf. Process. Syst. (NIPS)}}, Vol.~\bibinfo{volume}{2017-Decem}. \bibinfo{publisher}{Neural information processing systems foundation}, \bibinfo{pages}{1025--1035}.
\newblock


\bibitem[Hamouda et~al\mbox{.}(2011)]%
        {hamouda_reader_2011}
\bibfield{author}{\bibinfo{person}{Essia Hamouda}, \bibinfo{person}{Nathalie Mitton}, {and} \bibinfo{person}{David Simplot-Ryl}.} \bibinfo{year}{2011}\natexlab{}.
\newblock \showarticletitle{Reader {Anti}-collision in dense {RFID} networks with mobile tags}. In \bibinfo{booktitle}{\emph{2011 {IEEE} {International} {Conference} on {RFID}-{Technologies} and {Applications}}}. \bibinfo{pages}{327--334}.
\newblock
\urldef\tempurl%
\url{https://doi.org/10.1109/RFID-TA.2011.6068657}
\showDOI{\tempurl}


\bibitem[Huang et~al\mbox{.}(2024)]%
        {huang2024spe}
\bibfield{author}{\bibinfo{person}{Yinan Huang}, \bibinfo{person}{William Lu}, \bibinfo{person}{Joshua Robinson}, \bibinfo{person}{Yu Yang}, \bibinfo{person}{Muhan Zhang}, \bibinfo{person}{Stefanie Jegelka}, {and} \bibinfo{person}{Pan Li}.} \bibinfo{year}{2024}\natexlab{}.
\newblock \showarticletitle{On the Stability of Expressive Positional Encodings for Graph Neural Networks}. In \bibinfo{booktitle}{\emph{Proc. 12th International Conference on Learning Representations (ICLR'24)}}.
\newblock
\urldef\tempurl%
\url{https://openreview.net/forum?id=xAqcJ9XoTf}
\showURL{%
\tempurl}


\bibitem[{IEEE}(2016)]%
        {ieee802_15_4_2016}
\bibfield{author}{\bibinfo{person}{{IEEE}}.} \bibinfo{year}{2016}\natexlab{}.
\newblock \bibinfo{booktitle}{\emph{{IEEE} {Standard} for {Low}-{Rate} {Wireless} {Networks} –{Amendment} 2: {Ultra}-{Low} {Power} {Physical} {Layer}}}.
\newblock


\bibitem[{Iyer et al.}(2016)]%
        {iyer_inter-technology_2016}
\bibfield{author}{\bibinfo{person}{Vikram {Iyer et al.}}} \bibinfo{year}{2016}\natexlab{}.
\newblock \showarticletitle{Inter-{Technology} {Backscatter}: {Towards} {Internet} {Connectivity} for {Implanted} {Devices}}. \bibinfo{publisher}{ACM}, \bibinfo{pages}{356--369}.
\newblock
\showISBNx{978-1-4503-4193-6}
\urldef\tempurl%
\url{https://doi.org/10.1145/2934872.2934894}
\showDOI{\tempurl}


\bibitem[Jeon et~al\mbox{.}(2022)]%
        {jeon2022neural}
\bibfield{author}{\bibinfo{person}{Wonseok Jeon}, \bibinfo{person}{Mukul Gagrani}, \bibinfo{person}{Burak Bartan}, \bibinfo{person}{Weiliang~Will Zeng}, \bibinfo{person}{Harris Teague}, \bibinfo{person}{Piero Zappi}, {and} \bibinfo{person}{Christopher Lott}.} \bibinfo{year}{2022}\natexlab{}.
\newblock \showarticletitle{Neural DAG scheduling via one-shot priority sampling}. In \bibinfo{booktitle}{\emph{The Eleventh International Conference on Learning Representations}}.
\newblock


\bibitem[{Karimi} et~al\mbox{.}(2017)]%
        {karimi_design_2017}
\bibfield{author}{\bibinfo{person}{Y. {Karimi}}, \bibinfo{person}{A. {Athalye}}, \bibinfo{person}{S.~R. {Das}}, \bibinfo{person}{P.~M. {Djurić}}, {and} \bibinfo{person}{M. {Stanaćević}}.} \bibinfo{year}{2017}\natexlab{}.
\newblock \showarticletitle{Design of a backscatter-based Tag-to-Tag system}. In \bibinfo{booktitle}{\emph{2017 IEEE International Conference on RFID (IEEE RFID)}}. \bibinfo{pages}{6--12}.
\newblock
\urldef\tempurl%
\url{https://doi.org/10.1109/RFID.2017.7945579}
\showDOI{\tempurl}


\bibitem[Katanbaf et~al\mbox{.}(2021)]%
        {katanbaf2021multiscatter}
\bibfield{author}{\bibinfo{person}{Mohamad Katanbaf}, \bibinfo{person}{Ali Saffari}, {and} \bibinfo{person}{Joshua~R Smith}.} \bibinfo{year}{2021}\natexlab{}.
\newblock \showarticletitle{Multiscatter: Multistatic backscatter networking for battery-free sensors}. In \bibinfo{booktitle}{\emph{Proceedings of the 19th ACM Conference on Embedded Networked Sensor Systems}}. \bibinfo{pages}{69--83}.
\newblock


\bibitem[{Kellogg et al.}(2014)]%
        {kellogg2014wi}
\bibfield{author}{\bibinfo{person}{Bryce {Kellogg et al.}}} \bibinfo{year}{2014}\natexlab{}.
\newblock \showarticletitle{{Wi-Fi} {Backscatter}: {Internet} {Connectivity} for {RF}-powered {Devices}}. In \bibinfo{booktitle}{\emph{Proc. Special Interest Group Data Commun. (SIGCOMM)}}. \bibinfo{publisher}{ACM}, \bibinfo{address}{New York, NY, USA}, \bibinfo{pages}{607--618}.
\newblock
\showISBNx{978-1-4503-2836-4}
\urldef\tempurl%
\url{https://doi.org/10.1145/2619239.2626319}
\showDOI{\tempurl}


\bibitem[{Kellogg et al.}(2016)]%
        {kellogg_passive_2016}
\bibfield{author}{\bibinfo{person}{Bryce {Kellogg et al.}}} \bibinfo{year}{2016}\natexlab{}.
\newblock \showarticletitle{Passive {Wi}-{Fi}: {Bringing} {Low} {Power} to {Wi}-{Fi} {Transmissions}}. In \bibinfo{booktitle}{\emph{Proc. Symp. Networked Syst. Des. Implementation (NSDI)}}. \bibinfo{publisher}{NSDI}, \bibinfo{pages}{151--164}.
\newblock
\showISBNx{978-1-931971-29-4}


\bibitem[Kingma and {Lei Ba}(2015)]%
        {Kingma2015adam}
\bibfield{author}{\bibinfo{person}{Diederik~P Kingma} {and} \bibinfo{person}{Jimmy {Lei Ba}}.} \bibinfo{year}{2015}\natexlab{}.
\newblock \showarticletitle{{Adam}: A Method For Stochastic Optimization}. In \bibinfo{booktitle}{\emph{Proc. Int. Conf. Learn. Representations (ICLR)}}.
\newblock
\showeprint[arxiv]{1412.6980v9}


\bibitem[Kipf and Welling(2017)]%
        {Kipf2017gcn}
\bibfield{author}{\bibinfo{person}{Thomas~N. Kipf} {and} \bibinfo{person}{Max Welling}.} \bibinfo{year}{2017}\natexlab{}.
\newblock \showarticletitle{{Semi-Supervised Classification with Graph Convolutional Networks}}. In \bibinfo{booktitle}{\emph{Proc. 5th Int. Conf. Learn. Representations (ICLR)}}. \bibinfo{publisher}{ICLR}.
\newblock
\showeprint[arxiv]{1609.02907}


\bibitem[Kreuzer et~al\mbox{.}(2021)]%
        {kreuzer2021rethinking}
\bibfield{author}{\bibinfo{person}{Devin Kreuzer}, \bibinfo{person}{Dominique Beaini}, \bibinfo{person}{Will Hamilton}, \bibinfo{person}{Vincent L\'{e}tourneau}, {and} \bibinfo{person}{Prudencio Tossou}.} \bibinfo{year}{2021}\natexlab{}.
\newblock \showarticletitle{Rethinking Graph Transformers with Spectral Attention}. In \bibinfo{booktitle}{\emph{Advances in Neural Information Processing Systems}}, \bibfield{editor}{\bibinfo{person}{M.~Ranzato}, \bibinfo{person}{A.~Beygelzimer}, \bibinfo{person}{Y.~Dauphin}, \bibinfo{person}{P.S. Liang}, {and} \bibinfo{person}{J.~Wortman Vaughan}} (Eds.), Vol.~\bibinfo{volume}{34}. \bibinfo{publisher}{Curran Associates, Inc.}, \bibinfo{pages}{21618--21629}.
\newblock
\urldef\tempurl%
\url{https://proceedings.neurips.cc/paper_files/paper/2021/file/b4fd1d2cb085390fbbadae65e07876a7-Paper.pdf}
\showURL{%
\tempurl}


\bibitem[Krogh and Hertz(1991)]%
        {krogh1991simple}
\bibfield{author}{\bibinfo{person}{Anders Krogh} {and} \bibinfo{person}{John Hertz}.} \bibinfo{year}{1991}\natexlab{}.
\newblock \showarticletitle{A simple weight decay can improve generalization}.
\newblock \bibinfo{journal}{\emph{Advances in neural information processing systems}}  \bibinfo{volume}{4} (\bibinfo{year}{1991}).
\newblock


\bibitem[Kwak and Hong(2004)]%
        {kwak2004linear}
\bibfield{author}{\bibinfo{person}{Jin~Ho Kwak} {and} \bibinfo{person}{Sungpyo Hong}.} \bibinfo{year}{2004}\natexlab{}.
\newblock \bibinfo{booktitle}{\emph{Linear algebra}}.
\newblock \bibinfo{publisher}{Springer Science \& Business Media}.
\newblock


\bibitem[LeCun et~al\mbox{.}(1989)]%
        {lecun1989optimal}
\bibfield{author}{\bibinfo{person}{Yann LeCun}, \bibinfo{person}{John Denker}, {and} \bibinfo{person}{Sara Solla}.} \bibinfo{year}{1989}\natexlab{}.
\newblock \showarticletitle{Optimal brain damage}.
\newblock \bibinfo{journal}{\emph{Advances in neural information processing systems}}  \bibinfo{volume}{2} (\bibinfo{year}{1989}).
\newblock


\bibitem[Li et~al\mbox{.}(2018b)]%
        {li2018passive}
\bibfield{author}{\bibinfo{person}{Yan Li}, \bibinfo{person}{Zicheng Chi}, \bibinfo{person}{Xin Liu}, {and} \bibinfo{person}{Ting Zhu}.} \bibinfo{year}{2018}\natexlab{b}.
\newblock \showarticletitle{Passive-zigbee: Enabling zigbee communication in iot networks with 1000x+ less power consumption}. In \bibinfo{booktitle}{\emph{Proceedings of the 16th ACM conference on embedded networked sensor systems}}. \bibinfo{pages}{159--171}.
\newblock


\bibitem[Li et~al\mbox{.}(2018a)]%
        {Li2018copsgcn}
\bibfield{author}{\bibinfo{person}{Zhuwen Li}, \bibinfo{person}{Qifeng Chen}, {and} \bibinfo{person}{Vladlen Koltun}.} \bibinfo{year}{2018}\natexlab{a}.
\newblock \showarticletitle{Combinatorial optimization with graph convolutional networks and guided tree search}. In \bibinfo{booktitle}{\emph{Proc. Advances in Neural Inf. Process. Syst. (NeurIPS)}}. \bibinfo{pages}{539--548}.
\newblock


\bibitem[Lim et~al\mbox{.}(2023)]%
        {lim2022sign}
\bibfield{author}{\bibinfo{person}{Derek Lim}, \bibinfo{person}{Joshua Robinson}, \bibinfo{person}{Lingxiao Zhao}, \bibinfo{person}{Tess Smidt}, \bibinfo{person}{Suvrit Sra}, \bibinfo{person}{Haggai Maron}, {and} \bibinfo{person}{Stefanie Jegelka}.} \bibinfo{year}{2023}\natexlab{}.
\newblock \showarticletitle{Sign and basis invariant networks for spectral graph representation learning}.
\newblock  (\bibinfo{year}{2023}).
\newblock


\bibitem[Ma and Yarats(2021)]%
        {ma2021adequacy}
\bibfield{author}{\bibinfo{person}{Jerry Ma} {and} \bibinfo{person}{Denis Yarats}.} \bibinfo{year}{2021}\natexlab{}.
\newblock \showarticletitle{On the adequacy of untuned warmup for adaptive optimization}. In \bibinfo{booktitle}{\emph{Proceedings of the AAAI Conference on Artificial Intelligence}}, Vol.~\bibinfo{volume}{35}. \bibinfo{pages}{8828--8836}.
\newblock


\bibitem[{Majid} et~al\mbox{.}(2019)]%
        {jansen_multihopbackscatter_2019}
\bibfield{author}{\bibinfo{person}{A.~Y. {Majid}}, \bibinfo{person}{M. {Jansen}}, \bibinfo{person}{G.~O. {Delgado}}, \bibinfo{person}{K.~S. {Yildirim}}, {and} \bibinfo{person}{P. {Pawełłzak}}.} \bibinfo{year}{2019}\natexlab{}.
\newblock \showarticletitle{Multi-hop Backscatter Tag-to-Tag Networks}. In \bibinfo{booktitle}{\emph{Proc. Int. Conf. Comput. Commun. (INFOCOM)}}. \bibinfo{publisher}{IEEE}, \bibinfo{pages}{721--729}.
\newblock
\showISSN{2641-9874}
\urldef\tempurl%
\url{https://doi.org/10.1109/INFOCOM.2019.8737551}
\showDOI{\tempurl}


\bibitem[Manchanda et~al\mbox{.}(2020)]%
        {Manchanda2020learning}
\bibfield{author}{\bibinfo{person}{Sahil Manchanda}, \bibinfo{person}{Akash Mittal}, \bibinfo{person}{Anuj Dhawan}, \bibinfo{person}{Sourav Medya}, \bibinfo{person}{Sayan Ranu}, {and} \bibinfo{person}{Ambuj Singh}.} \bibinfo{year}{2020}\natexlab{}.
\newblock \showarticletitle{{Learning Heuristics over Large Graphs via Deep Reinforcement Learning}}. In \bibinfo{booktitle}{\emph{Proc. 34th Conf. Neural Inf. Process. Syst. (NIPS)}}.
\newblock
\showeprint[arxiv]{1903.03332}


\bibitem[Mao et~al\mbox{.}(2019)]%
        {mao2019learning}
\bibfield{author}{\bibinfo{person}{Hongzi Mao}, \bibinfo{person}{Malte Schwarzkopf}, \bibinfo{person}{Shaileshh~Bojja Venkatakrishnan}, \bibinfo{person}{Zili Meng}, {and} \bibinfo{person}{Mohammad Alizadeh}.} \bibinfo{year}{2019}\natexlab{}.
\newblock \showarticletitle{Learning scheduling algorithms for data processing clusters}.
\newblock In \bibinfo{booktitle}{\emph{Proceedings of the ACM special interest group on data communication}}. \bibinfo{pages}{270--288}.
\newblock


\bibitem[Nakkiran et~al\mbox{.}(2020)]%
        {Nakkiran2020DeepDD}
\bibfield{author}{\bibinfo{person}{Preetum Nakkiran}, \bibinfo{person}{Gal Kaplun}, \bibinfo{person}{Yamini Bansal}, \bibinfo{person}{Tristan Yang}, \bibinfo{person}{Boaz Barak}, {and} \bibinfo{person}{Ilya Sutskever}.} \bibinfo{year}{2020}\natexlab{}.
\newblock \showarticletitle{Deep Double Descent: Where Bigger Models and More Data Hurt}. In \bibinfo{booktitle}{\emph{International Conference on Learning Representations (ICLR)}}.
\newblock
\urldef\tempurl%
\url{https://openreview.net/forum?id=B1g5sA4twr}
\showURL{%
\tempurl}


\bibitem[Oikonomou et~al\mbox{.}(2022)]%
        {oikonomou22contiking}
\bibfield{author}{\bibinfo{person}{George Oikonomou}, \bibinfo{person}{Simon Duquennoy}, \bibinfo{person}{Atis Elsts}, \bibinfo{person}{Joakim Eriksson}, \bibinfo{person}{Yasuyuki Tanaka}, {and} \bibinfo{person}{Nicolas Tsiftes}.} \bibinfo{year}{2022}\natexlab{}.
\newblock \showarticletitle{The {Contiki-NG} open source operating system for next generation {IoT} devices}.
\newblock \bibinfo{journal}{\emph{SoftwareX}}  \bibinfo{volume}{18} (\bibinfo{year}{2022}), \bibinfo{pages}{101089}.
\newblock
\showISSN{2352-7110}
\urldef\tempurl%
\url{https://doi.org/10.1016/j.softx.2022.101089}
\showDOI{\tempurl}


\bibitem[Paszke et~al\mbox{.}(2019)]%
        {pytorch2019}
\bibfield{author}{\bibinfo{person}{Adam Paszke}, \bibinfo{person}{Sam Gross}, \bibinfo{person}{Francisco Massa}, \bibinfo{person}{Adam Lerer}, \bibinfo{person}{James Bradbury}, \bibinfo{person}{Gregory Chanan}, {and} \bibinfo{person}{et al.}} \bibinfo{year}{2019}\natexlab{}.
\newblock \showarticletitle{PyTorch: An Imperative Style, High-Performance Deep Learning Library}. In \bibinfo{booktitle}{\emph{Proc. Advances Neural Inf. Process. Syst.}}, \bibfield{editor}{\bibinfo{person}{H.~Wallach}, \bibinfo{person}{H.~Larochelle}, \bibinfo{person}{A.~Beygelzimer}, \bibinfo{person}{F.~d\textquotesingle Alch\'{e}-Buc}, \bibinfo{person}{E.~Fox}, {and} \bibinfo{person}{R.~Garnett}} (Eds.). \bibinfo{publisher}{Curran Associates, Inc.}, \bibinfo{pages}{8024--8035}.
\newblock


\bibitem[P{\'{e}}rez-Penichet et~al\mbox{.}(2016)]%
        {Perez-Penichet2016augmenting}
\bibfield{author}{\bibinfo{person}{Carlos P{\'{e}}rez-Penichet}, \bibinfo{person}{Frederik Hermans}, \bibinfo{person}{Ambuj Varshney}, {and} \bibinfo{person}{Thiemo Voigt}.} \bibinfo{year}{2016}\natexlab{}.
\newblock \showarticletitle{{Augmenting IoT networks with backscatter-enabled passive sensor tags}}. In \bibinfo{booktitle}{\emph{Proc. Annu. Int. Conf. Mobile Comput. Netw. (MOBICOM)}}. \bibinfo{publisher}{ACM}, \bibinfo{pages}{23--27}.
\newblock
\showISBNx{9781450342513}
\urldef\tempurl%
\url{https://doi.org/10.1145/2980115.2980132}
\showDOI{\tempurl}


\bibitem[P{\'{e}}rez-Penichet et~al\mbox{.}(2020)]%
        {PerezPenichet2020afast}
\bibfield{author}{\bibinfo{person}{Carlos P{\'{e}}rez-Penichet}, \bibinfo{person}{Dilushi Piumwardane}, \bibinfo{person}{Christian Rohner}, {and} \bibinfo{person}{Thiemo Voigt}.} \bibinfo{year}{2020}\natexlab{}.
\newblock \showarticletitle{{A Fast Carrier Scheduling Algorithm for Battery-free Sensor Tags in Commodity Wireless Networks}}. In \bibinfo{booktitle}{\emph{Proc. Int. Conf. Comput. Commun. (INFOCOM)}}. \bibinfo{publisher}{IEEE}, \bibinfo{pages}{994--1003}.
\newblock
\showISBNx{9781728164120}
\showISSN{0743166X}
\urldef\tempurl%
\url{https://doi.org/10.1109/infocom41043.2020.9155241}
\showDOI{\tempurl}


\bibitem[Perez-Ramirez et~al\mbox{.}(2023)]%
        {perezramirez_DeepGANTT_2023}
\bibfield{author}{\bibinfo{person}{Daniel~F. Perez-Ramirez}, \bibinfo{person}{Carlos P\'{e}rez-Penichet}, \bibinfo{person}{Nicolas Tsiftes}, \bibinfo{person}{Thiemo Voigt}, \bibinfo{person}{Dejan Kosti\'{c}}, {and} \bibinfo{person}{Magnus Boman}.} \bibinfo{year}{2023}\natexlab{}.
\newblock \showarticletitle{DeepGANTT: A Scalable Deep Learning Scheduler for Backscatter Networks}. In \bibinfo{booktitle}{\emph{Proceedings of the 22nd International Conference on Information Processing in Sensor Networks}} (San Antonio, TX, USA) \emph{(\bibinfo{series}{IPSN '23})}. \bibinfo{publisher}{Association for Computing Machinery}, \bibinfo{address}{New York, NY, USA}, \bibinfo{pages}{163–176}.
\newblock
\showISBNx{9798400701184}
\urldef\tempurl%
\url{https://doi.org/10.1145/3583120.3586957}
\showDOI{\tempurl}


\bibitem[Pérez-Penichet et~al\mbox{.}(2020)]%
        {perez-penichet_tagalong_2020}
\bibfield{author}{\bibinfo{person}{Carlos Pérez-Penichet}, \bibinfo{person}{Dilushi Piumwardane}, \bibinfo{person}{Christian Rohner}, {and} \bibinfo{person}{Thiemo Voigt}.} \bibinfo{year}{2020}\natexlab{}.
\newblock \showarticletitle{{TagAlong}: {E}fficient {I}ntegration of {B}attery-{F}ree {S}ensor {T}ags in {S}tandard {W}ireless {N}etworks}. In \bibinfo{booktitle}{\emph{Proc. 19th ACM/IEEE Int. Conf. Inf. Process. Sensor Netw. (IPSN)}}. \bibinfo{address}{Sydney, Australia}.
\newblock
\urldef\tempurl%
\url{https://doi.org/10.1109/IPSN48710.2020.00020}
\showDOI{\tempurl}


\bibitem[Radford et~al\mbox{.}(2018)]%
        {radford2018gptimproving}
\bibfield{author}{\bibinfo{person}{Alec Radford}, \bibinfo{person}{Karthik Narasimhan}, \bibinfo{person}{Tim Salimans}, \bibinfo{person}{Ilya Sutskever}, {et~al\mbox{.}}} \bibinfo{year}{2018}\natexlab{}.
\newblock \showarticletitle{Improving language understanding by generative pre-training}.
\newblock  (\bibinfo{year}{2018}).
\newblock


\bibitem[Radford et~al\mbox{.}(2019)]%
        {radford2019gptlanguage}
\bibfield{author}{\bibinfo{person}{Alec Radford}, \bibinfo{person}{Jeffrey Wu}, \bibinfo{person}{Rewon Child}, \bibinfo{person}{David Luan}, \bibinfo{person}{Dario Amodei}, \bibinfo{person}{Ilya Sutskever}, {et~al\mbox{.}}} \bibinfo{year}{2019}\natexlab{}.
\newblock \showarticletitle{Language models are unsupervised multitask learners}.
\newblock \bibinfo{journal}{\emph{OpenAI blog}} \bibinfo{volume}{1}, \bibinfo{number}{8} (\bibinfo{year}{2019}), \bibinfo{pages}{9}.
\newblock


\bibitem[Ramp{\'a}{\v{s}}ek et~al\mbox{.}(2022)]%
        {rampavsek2022recipe}
\bibfield{author}{\bibinfo{person}{Ladislav Ramp{\'a}{\v{s}}ek}, \bibinfo{person}{Michael Galkin}, \bibinfo{person}{Vijay~Prakash Dwivedi}, \bibinfo{person}{Anh~Tuan Luu}, \bibinfo{person}{Guy Wolf}, {and} \bibinfo{person}{Dominique Beaini}.} \bibinfo{year}{2022}\natexlab{}.
\newblock \showarticletitle{Recipe for a general, powerful, scalable graph transformer}.
\newblock \bibinfo{journal}{\emph{Advances in Neural Information Processing Systems}}  \bibinfo{volume}{35} (\bibinfo{year}{2022}), \bibinfo{pages}{14501--14515}.
\newblock


\bibitem[Scarselli et~al\mbox{.}(2009)]%
        {Scarselli2009gnnmodel}
\bibfield{author}{\bibinfo{person}{Franco Scarselli}, \bibinfo{person}{Marco Gori}, \bibinfo{person}{Ah~Chung Tsoi}, \bibinfo{person}{Markus Hagenbuchner}, {and} \bibinfo{person}{Gabriele Monfardini}.} \bibinfo{year}{2009}\natexlab{}.
\newblock \showarticletitle{{The graph neural network model}}.
\newblock \bibinfo{journal}{\emph{IEEE Trans. Neural Netw.}} \bibinfo{volume}{20}, \bibinfo{number}{1} (\bibinfo{date}{jan} \bibinfo{year}{2009}), \bibinfo{pages}{61--80}.
\newblock
\showISSN{10459227}
\urldef\tempurl%
\url{https://doi.org/10.1109/TNN.2008.2005605}
\showDOI{\tempurl}


\bibitem[Talla et~al\mbox{.}(2017)]%
        {talla2017lora}
\bibfield{author}{\bibinfo{person}{Vamsi Talla}, \bibinfo{person}{Mehrdad Hessar}, \bibinfo{person}{Bryce Kellogg}, \bibinfo{person}{Ali Najafi}, \bibinfo{person}{Joshua~R. Smith}, {and} \bibinfo{person}{Shyamnath Gollakota}.} \bibinfo{year}{2017}\natexlab{}.
\newblock \showarticletitle{{LoRa} {Backscatter}: {Enabling} {The} {Vision} of {Ubiquitous} {Connectivity}}.
\newblock \bibinfo{journal}{\emph{Proc. ACM Interact. Mob. Wearable Ubiquitous Technol.}} \bibinfo{volume}{1}, \bibinfo{number}{3}, \bibinfo{pages}{105:1--105:24}.
\newblock
\showISSN{2474-9567}
\urldef\tempurl%
\url{https://doi.org/10.1145/3130970}
\showDOI{\tempurl}


\bibitem[Vaswani et~al\mbox{.}(2017)]%
        {Vaswani2017attention}
\bibfield{author}{\bibinfo{person}{Ashish Vaswani}, \bibinfo{person}{Noam Shazeer}, \bibinfo{person}{Niki Parmar}, \bibinfo{person}{Jakob Uszkoreit}, \bibinfo{person}{Llion Jones}, \bibinfo{person}{Aidan~N. Gomez}, \bibinfo{person}{{\L}ukasz Kaiser}, {and} \bibinfo{person}{Illia Polosukhin}.} \bibinfo{year}{2017}\natexlab{}.
\newblock \showarticletitle{{Attention is all you need}}. In \bibinfo{booktitle}{\emph{Proc. Advances Neural Inf. Process. Syst. (NIPS)}}, Vol.~\bibinfo{volume}{2017-Decem}. \bibinfo{publisher}{NIPS}, \bibinfo{pages}{5999--6009}.
\newblock
\showISSN{10495258}


\bibitem[Veli{\v{c}}kovi{\'{c}} et~al\mbox{.}(2018)]%
        {Velickovic2018gat}
\bibfield{author}{\bibinfo{person}{Petar Veli{\v{c}}kovi{\'{c}}}, \bibinfo{person}{Arantxa Casanova}, \bibinfo{person}{Pietro Li{\`{o}}}, \bibinfo{person}{Guillem Cucurull}, \bibinfo{person}{Adriana Romero}, {and} \bibinfo{person}{Yoshua Bengio}.} \bibinfo{year}{2018}\natexlab{}.
\newblock \showarticletitle{{Graph attention networks}}. In \bibinfo{booktitle}{\emph{Proc. 6th Int. Conf. Learn. Representations (ICLR)}}. \bibinfo{publisher}{ICLR}.
\newblock
\showeprint[arxiv]{1710.10903}


\bibitem[Vesselinova et~al\mbox{.}(2020)]%
        {vesselinova2020learning}
\bibfield{author}{\bibinfo{person}{Natalia Vesselinova}, \bibinfo{person}{Rebecca Steinert}, \bibinfo{person}{Daniel~F Perez-Ramirez}, {and} \bibinfo{person}{Magnus Boman}.} \bibinfo{year}{2020}\natexlab{}.
\newblock \showarticletitle{Learning combinatorial optimization on graphs: A survey with applications to networking}.
\newblock \bibinfo{journal}{\emph{IEEE Access}}  \bibinfo{volume}{8} (\bibinfo{year}{2020}), \bibinfo{pages}{120388--120416}.
\newblock


\bibitem[Vinyals et~al\mbox{.}(2015)]%
        {Vinyals2015pointernets}
\bibfield{author}{\bibinfo{person}{Oriol Vinyals}, \bibinfo{person}{Google Brain}, \bibinfo{person}{Meire Fortunato}, {and} \bibinfo{person}{Navdeep Jaitly}.} \bibinfo{year}{2015}\natexlab{}.
\newblock \showarticletitle{{Pointer Networks}}. In \bibinfo{booktitle}{\emph{Proc. Advances Neural Inf. Process. Syst. (NIPS)}}. \bibinfo{pages}{2692--2700}.
\newblock


\bibitem[Wang et~al\mbox{.}(2022)]%
        {wang2022equivariant}
\bibfield{author}{\bibinfo{person}{Haorui Wang}, \bibinfo{person}{Haoteng Yin}, \bibinfo{person}{Muhan Zhang}, {and} \bibinfo{person}{Pan Li}.} \bibinfo{year}{2022}\natexlab{}.
\newblock \showarticletitle{Equivariant and stable positional encoding for more powerful graph neural networks}. In \bibinfo{booktitle}{\emph{Proc. 10th International Conference on Learning Representations (ICLR'22)}}.
\newblock


\bibitem[Winter(2012)]%
        {winter2012rpl}
\bibfield{author}{\bibinfo{person}{T. Winter}.} \bibinfo{year}{2012}\natexlab{}.
\newblock \bibinfo{title}{RPL: IPv6 Routing Protocol for Low-Power and Lossy Networks}.
\newblock
\newblock
\urldef\tempurl%
\url{https://www.rfc-editor.org/rfc/rfc6550}
\showURL{%
Retrieved Oct. 2022 from \tempurl}


\bibitem[Wu et~al\mbox{.}(2021)]%
        {zonghan2021gnnsurvey}
\bibfield{author}{\bibinfo{person}{Zonghan Wu}, \bibinfo{person}{Shirui Pan}, \bibinfo{person}{Fengwen Chen}, \bibinfo{person}{Guodong Long}, \bibinfo{person}{Chengqi Zhang}, {and} \bibinfo{person}{Philip~S. Yu}.} \bibinfo{year}{2021}\natexlab{}.
\newblock \showarticletitle{A Comprehensive Survey on Graph Neural Networks}.
\newblock \bibinfo{journal}{\emph{{IEEE} Trans. Neural Netw.}} \bibinfo{volume}{32}, \bibinfo{number}{1} (\bibinfo{year}{2021}), \bibinfo{pages}{4--24}.
\newblock
\urldef\tempurl%
\url{https://doi.org/10.1109/TNNLS.2020.2978386}
\showDOI{\tempurl}


\bibitem[Xu et~al\mbox{.}(2019)]%
        {xu2018powerful}
\bibfield{author}{\bibinfo{person}{Keyulu Xu}, \bibinfo{person}{Weihua Hu}, \bibinfo{person}{Jure Leskovec}, {and} \bibinfo{person}{Stefanie Jegelka}.} \bibinfo{year}{2019}\natexlab{}.
\newblock \showarticletitle{How powerful are graph neural networks?}. In \bibinfo{booktitle}{\emph{Proc. Int. Conf. Learn. Representations (ICLR'19)}}.
\newblock


\bibitem[{Yang} et~al\mbox{.}(2011)]%
        {liu_season_2011}
\bibfield{author}{\bibinfo{person}{L. {Yang}}, \bibinfo{person}{J. {Han}}, \bibinfo{person}{Y. {Qi}}, \bibinfo{person}{C. {Wang}}, \bibinfo{person}{T. {Gu}}, {and} \bibinfo{person}{Y. {Liu}}.} \bibinfo{year}{2011}\natexlab{}.
\newblock \showarticletitle{Season: Shelving interference and joint identification in large-scale {RFID} systems}. In \bibinfo{booktitle}{\emph{Proc. Int. Conf. Comput. Commun. (INFOCOM)}}. \bibinfo{publisher}{IEEE}, \bibinfo{pages}{3092--3100}.
\newblock
\showISSN{0743-166X}
\urldef\tempurl%
\url{https://doi.org/10.1109/INFCOM.2011.5935154}
\showDOI{\tempurl}


\bibitem[{Yue} et~al\mbox{.}(2012)]%
        {chen_time-efficient_2012}
\bibfield{author}{\bibinfo{person}{H. {Yue}}, \bibinfo{person}{C. {Zhang}}, \bibinfo{person}{M. {Pan}}, \bibinfo{person}{Y. {Fang}}, {and} \bibinfo{person}{S. {Chen}}.} \bibinfo{year}{2012}\natexlab{}.
\newblock \showarticletitle{A time-efficient information collection protocol for large-scale {RFID} systems}. In \bibinfo{booktitle}{\emph{Proc. Int. Conf. Comput. Commun. (INFOCOM)}}. \bibinfo{publisher}{IEEE}, \bibinfo{pages}{2158--2166}.
\newblock
\showISSN{0743-166X}
\urldef\tempurl%
\url{https://doi.org/10.1109/INFCOM.2012.6195599}
\showDOI{\tempurl}


\bibitem[Zhang et~al\mbox{.}(2017)]%
        {zhang2017freerider}
\bibfield{author}{\bibinfo{person}{Pengyu Zhang}, \bibinfo{person}{Colleen Josephson}, \bibinfo{person}{Dinesh Bharadia}, {and} \bibinfo{person}{Sachin Katti}.} \bibinfo{year}{2017}\natexlab{}.
\newblock \showarticletitle{Freerider: Backscatter communication using commodity radios}. In \bibinfo{booktitle}{\emph{Proceedings of the 13th international conference on emerging networking experiments and technologies}}. \bibinfo{pages}{389--401}.
\newblock


\end{thebibliography}

\acrodef{TSCH}{Time-Slotted Channel Hopping}
\acrodef{RF}{Radio Frequency}
\acrodefindefinite{RF}{an}{a}
\acrodef{LNA}{Low-Noise Amplifier}
\acrodefindefinite{LNA}{an}{a}
\acrodef{LO}{Local Oscillator}
\acrodefindefinite{LO}{an}{a}
\acrodef{ADC}{Analog-to-Digital Converter}
\acrodef{IF}{Intermediate Frequency}
\acrodef{CMOS}{Complementary Metal-Oxide-Semiconductor}
\acrodef{DBP}{Digital Baseband Processor}
\acrodef{DSSS}{Direct Sequence Spread Spectrum}
\acrodef{ASK}{Amplitude Shift Keying}
\acrodef{MSK}{Minimum Shift Keying}
\acrodefindefinite{MSK}{an}{a}
\acrodef{FSK}{Frequency Shift Keying}
\acrodefindefinite{FSK}{an}{a}
\acrodef{oqpsk}[O-QPSK]{Offset-Quadrature Phase Shift Keying}
\acrodef{BPF}{Band-Pass Filter}
\acrodef{PRR}{Packet Reception Ratio}
\acrodef{SDR}{Software Defined Radio}
\acrodefindefinite{SDR}{an}{a}
\acrodef{IC}{Integrated Circuit}
\acrodef{WSN}{Wireless Sensor Networks}
\acrodef{CCA}{Clear Channel Assessment}
\acrodef{MAC}{Medium Access Control}
\acrodef{IoT}{Internet of Things}
\acrodef{COTS}{Commercial Off-The-Shelf}

\acrodef{CO}{Constraint Optimizer}
\acrodef{COP}{Combinatorial Optimization Problem}
\acrodef{EVD}{Eigenvalue Decomposition}
\acrodef{ML}{Machine Learning}
\acrodef{MLP}{Multilayer Perceptron}
\acrodef{PE}{Positional Encoding}
\acrodefindefinite{ML}{an}{a}
\acrodef{DL}{Deep Learning}
\acrodef{GRL}{Graph Representation Learning}
\acrodef{GNN}{Graph Neural Network}
\acrodef{BN}{Bayesian Network}
\acrodef{RNN}{Recurrent Neural Network}
\acrodefindefinite{RNN}{an}{a}
\acrodef{NMT}{Neural Machine Translation}
\acrodefindefinite{NMT}{an}{a}
\acrodef{RL}{Reinforcement Learning}
\acrodefindefinite{RL}{an}{a}
\acrodef{DRL}{Deep Reinforcement Learning}
\acrodef{SSL}{Self-Supervised Learning}
\acrodefindefinite{SSL}{an}{a}
\acrodef{seq2seq}{Sequence-to-Sequence}
\acrodef{LSTM}{Long Short-Term Memory}
\acrodefindefinite{LSTM}{an}{a}
\acrodef{KDE}{Kernel Density Estimator}

\end{document}